%% file: main.tex
\documentclass[lettersize,journal]{IEEEtran}
\usepackage{amsmath,amsfonts}
\usepackage{graphicx}
\usepackage{algorithmic}
\usepackage{array}
\usepackage[caption=false,font=normalsize,
labelfont=sf,textfont=sf]{subfig}
\usepackage{textcomp}
\usepackage{stfloats}
\usepackage{url}
\usepackage{verbatim}
\usepackage{graphicx}
\usepackage{balance}
\usepackage{cite}
\usepackage{pifont}
\usepackage{booktabs}
\usepackage{multirow}
\usepackage{xspace}
\usepackage{multirow}
\usepackage{float}
\usepackage[table]{xcolor}
\definecolor{Gray}{gray}{0.9}
\definecolor{LightCyan}{rgb}{0.88,0.95,1}
\definecolor{blond}{rgb}{0.98, 0.94, 0.75}
\definecolor{LightGreen}{rgb}{0.333, 0.937, 0.769}
\definecolor{cross-attn-blue}{HTML}{449fd9}
\definecolor{cross-attn-yellow}{HTML}{f59a1a}
\usepackage{hyperref}
\usepackage[font=small,labelfont=bf]{caption}
\usepackage{subfig}
\usepackage{tcolorbox}
\usepackage{enumitem}

\def \ie {\emph{i.e.}}
\def \eg {\emph{e.g.}}
\def \etal {\emph{et al.}}

\newcommand{\tit}[1]{\smallbreak\noindent\textbf{#1.}}
\newcommand{\tinytit}[1]{\noindent\textbf{#1.}}

\newcommand{\cmark}{\ding{51}}
\newcommand{\xmark}{\ding{55}}

\newcommand{\ours}{Ti-MGD\xspace}

\newcommand{\dataset}{Dress Code Multimodal\xspace}
\newcommand{\datasetviton}{VITON-HD Multimodal\xspace}

\newcommand{\img}{I}
\newcommand{\pose}{P}
\newcommand{\latentpose}{p}
\newcommand{\sketch}{S}
\newcommand{\latentsketch}{s}
\newcommand{\textual}{Y}
\newcommand{\texture}{X}
\newcommand{\newimg}{\Tilde{I}}
\newcommand{\CLIPtextenc}{T_{E}}
\newcommand{\CLIPvisenc}{V_{E}}
\newcommand{\CLIPlookup}{E_{L}}
\newcommand{\CLIPtexttrans}{T_{T}}
\newcommand{\decoder}{\mathcal{D}}
\newcommand{\encoder}{\mathcal{E}}

\newcommand{\maskedimg}{I_M}
\newcommand{\latentmask}{m}
\newcommand{\mask}{M}

\newcommand{\tiadapter}{F_{\theta}}

\newcommand{\RNum}[1]{\lowercase\expandafter{\romannumeral #1\relax}}

\title{Multimodal-Conditioned Latent Diffusion Models\\for Fashion Image Editing}

\author{Alberto Baldrati$^*$, Davide Morelli$^*$, Marcella Cornia, Marco Bertini, Rita Cucchiara
\thanks{The first two authors equally contributed to this research.}
\thanks{A. Baldrati and M. Bertini are with the Media Integration and Communication Center (MICC), University of Florence, Italy  (e-mail: \{alberto.baldrati, marco.bertini\}@unifi.it).}
\thanks{D. Morelli and R. Cucchiara are with the Department of Engineering ``Enzo Ferrari'', University of Modena and Reggio Emilia, Italy (e-mail: \{davide.morelli, rita.cucchiara\}@unimore.it).}
\thanks{A. Baldrati and D. Morelli are also with the University of Pisa, Italy.}
\thanks{M. Cornia is with the Department of Education and Humanities, University of Modena and Reggio Emilia, Italy (e-mail: marcella.cornia@unimore.it).}
\thanks{Manuscript received March, 2024.}
}

\makeatletter
\let\@oldmaketitle\@maketitle
\renewcommand{\@maketitle}{\@oldmaketitle
\setcounter{figure}{0}
\includegraphics[width=\linewidth]{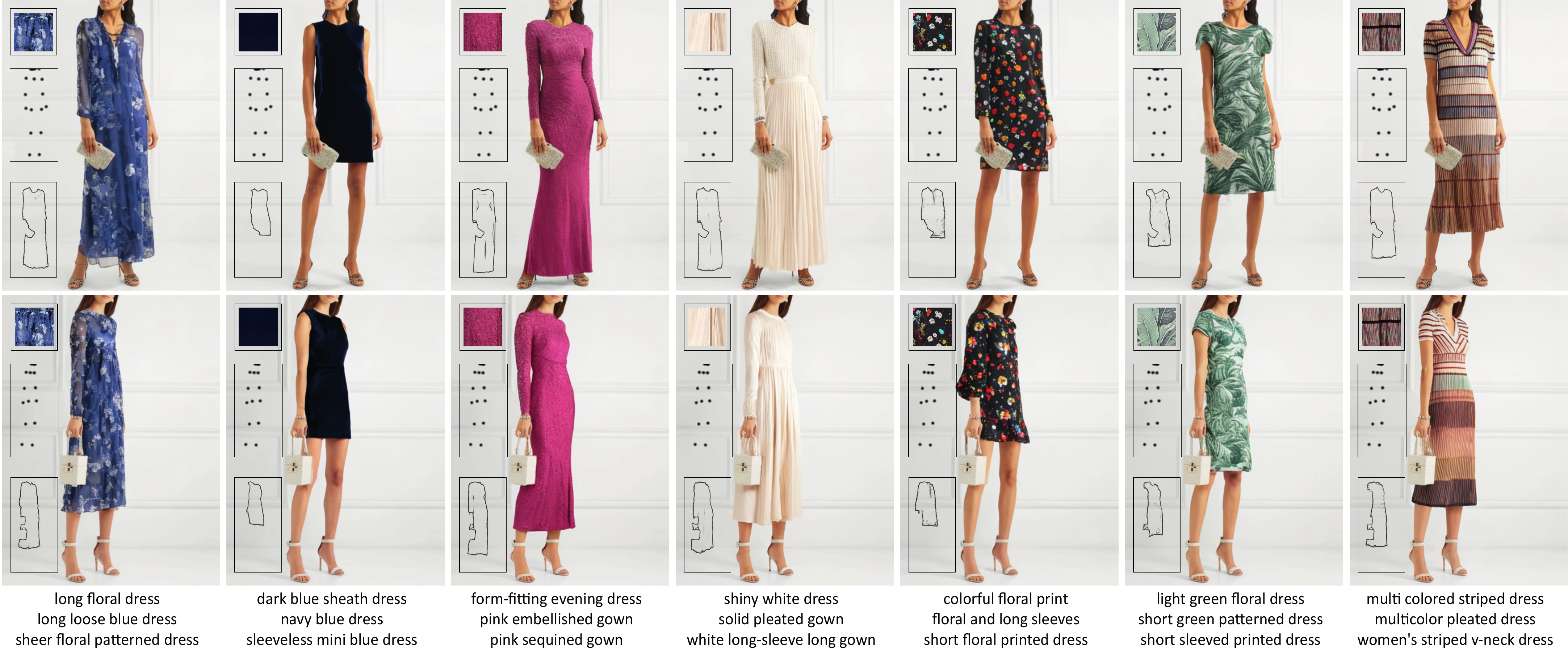}
\captionof{figure}{Example of images generated using the proposed Textual-inverted Multimodal Garment Designer (\ours) method, with each row featuring the same model edited using different inputs. For each generated image, we show the generation input conditions: texture (top left), keypoints (middle left), sketch (bottom left), and text (bottom of each column).}
\label{fig:qualitative-first-page}
\vspace{-4ex}}
\makeatother

\begin{document}
\setcounter{figure}{0}
\maketitle
\begin{abstract}
Fashion illustration is a crucial medium for designers to convey their creative vision and transform design concepts into tangible representations that showcase the interplay between clothing and the human body. In the context of fashion design, computer vision techniques have the potential to enhance and streamline the design process. Departing from prior research primarily focused on virtual try-on, this paper tackles the task of multimodal-conditioned fashion image editing. Our approach aims to generate human-centric fashion images guided by multimodal prompts, including text, human body poses, garment sketches, and fabric textures. To address this problem, we propose extending latent diffusion models to incorporate these multiple modalities and modifying the structure of the denoising network, taking multimodal prompts as input. To condition the proposed architecture on fabric textures, we employ textual inversion techniques and let diverse cross-attention layers of the denoising network attend to textual and texture information, thus incorporating different granularity conditioning details. Given the lack of datasets for the task, we extend two existing fashion datasets, Dress Code and VITON-HD, with multimodal annotations.
Experimental evaluations demonstrate the effectiveness of our proposed approach in terms of realism and coherence concerning the provided multimodal inputs.
\end{abstract}


\section{Introduction}
\label{sec:intro}
\input{sections/01_introduction.tex}

\section{Related Work}
\label{sec:related}
\input{sections/02_related.tex}

\section{Proposed Method}
\label{sec:method}
\input{sections/03_method.tex}

\section{Collecting Multimodal Fashion Datasets}
\label{sec:dataset}
\input{sections/04_dataset.tex}

\section{Experimental Evaluation}
\label{sec:experiments}
\input{sections/05_experiments.tex}

\section{Conclusion}
\label{sec:conclusion}
\input{sections/06_conclusion.tex}

\section*{Acknowledgments}
This work has been supported by the European Commission under the PNRR-M4C2 project ``FAIR - Future Artificial Intelligence Research'' and the European Horizon 2020 Programme (grant number 951911 - AI4Media), and by the PRIN project ``CREATIVE: CRoss-modal understanding and gEnerATIon of Visual and tExtual content'' (CUP B87G22000460001), co-funded by the Italian Ministry of University.

{\small
\bibliographystyle{IEEEtran}
\bibliography{bibliography}
}

\end{document}

%% file: sections/01_introduction.tex
\IEEEPARstart{I}{n} recent years, the intersection of computer vision and fashion has garnered significant attention, with a surge in research mainly dedicated to adapting or re-designing state-of-the-art computer vision models for fashion images. Previous studies have primarily focused on tasks such as clothing item recognition and retrieval~\cite{hadi2015buy,liu2016deepfashion,cartella2023openfashionclip, morelli2021fashionsearch++}, garment and outfit recommendation~\cite{hsiao2018creating,cucurull2019context,sarkar2023outfittransformer}, and virtual try-on~\cite{han2018viton,wang2018toward,yang2020towards,morelli2022dresscode,choi2021viton, morelli2023ladi}. While these works have advanced research in the field, limited attention has been paid to text-conditioned fashion image editing, mainly due to the specificity of the fashion lexicon, the lack of existing datasets, and the complexity of the task itself. Among the few works that have addressed the task, some attempts~\cite{zhu2017your,jiang2022text2human,pernuvs2023fice} have been dedicated to the use of GAN-based methods to generate images of models wearing clothing items only exploiting the condition of textual descriptions. 
Recently, diffusion models~\cite{ho2020denoising,dhariwal2021diffusion,nichol2021improved,rombach2022high} have shown exceptional generation capabilities compared to GANs, enabling better control over the synthesized output.
However, the applicability of these models to the fashion domain remains largely underexplored.

In this work, we go beyond standard text-conditioned generation and introduce \emph{multimodal-conditioned fashion image editing}, a new challenging task that involves the generation of new garment images worn by a given person, leveraging the conditioning of multiple multimodal constraints including human pose, garment sketches, textual descriptions, and garment fabric textures. The integration of multiple prompts in generative models is a complex computer vision task, especially in the context of garment images. In fact, these images exhibit considerable variation, influenced by factors such as the target gender (\ie~whether the garment is designed for men or women), the garment category, and target market dynamics (\ie~whether the garment is a luxury or economical item). At the same time, this task can have a significant impact on creative industries, as it can enable fashion designers to empower the design of new fashion items, facilitating the exploration of the interplay between their sketches, the available fabric textures, and diverse human body shapes.

To tackle the newly proposed task, we present a novel approach that enables the generative process to be guided by multimodal prompts (\ie~text, human pose, garment sketches, and fabric textures) while preserving the identity and body shape of the subject (Fig.~\ref{fig:qualitative-first-page}). Specifically, we leverage latent diffusion models~\cite{rombach2022high}, which define the forward and reverse processes in the latent space of a pre-trained autoencoder instead of the pixel space, and propose a denoising network that can be conditioned by multiple modalities, also incorporating pose consistency between input and generated images. A first attempt at fashion image editing conditioned by multimodal inputs has previously been proposed by us in~\cite{baldrati2023multimodal}. Compared to the previous version, we improve the architecture by enabling it to also deal with fabric texture input while retaining the capability to remove any constraint at inference time. In particular, we design a novel textual inversion-based component that can project texture images to the textual space of the diffusion model. We then let diverse cross-attention layers of the denoising network capture diverse granularity details, enabling simultaneous conditioning of both text and fabric textures through the same layers of the denoising network. We denote this new version as Textual-inverted Multimodal Garment Designer (\ours).

The task of multimodal-conditioned fashion image editing is new and no datasets are available both for training and testing. To effectively address the task, we also define a semi-automatic framework for extending existing fashion datasets with multimodal data. Specifically, we leverage two well-known virtual try-on datasets, Dress Code~\cite{morelli2022dresscode} and VITON-HD~\cite{choi2021viton}, and augment them with textual descriptions, garment sketches, and fabric texture. 

To evaluate the impact of conditioning signals, we introduce three novel evaluation metrics that measure human pose, sketch, and fabric texture coherence between input and generated images. Through extensive experiments on the proposed multimodal fashion benchmarks, we demonstrate the quantitative and qualitative effectiveness of our proposed approach in generating high-quality images based on multimodal inputs. 
As quantitative metrics and human evaluations confirm, our method outperforms state-of-the-art competitors and baselines.

In summary, our contributions are as follows: 
\begin{itemize}[noitemsep,topsep=0pt]
    \item We propose the novel task of multimodal-conditioned fashion image editing, which utilizes multimodal prompts to guide the generative process. 
    \item To tackle the task, we design a semi-automatic annotation framework to extend two existing fashion datasets with textual data, garment sketches, and fabric textures.
    \item We introduce a new human-centric generative architecture based on latent diffusion models capable of incorporating multimodal prompts while preserving the input person's characteristics. Specifically, we let the denoising network take multimodal prompts as input and design a novel textual inversion-based component that effectively integrates fabric texture by projecting texture images into the textual space of the diffusion model.
    \item To the best of our knowledge, we are the first to use, in a concrete working case, the property that distinct cross-attention layers of the denoising network can capture diverse granularity conditioning details. This method enables concurrent textual and texture generation conditioning by sharing the same layers. 
    \item Extensive experiments demonstrate that our proposed approach outperforms state-of-the-art competitors in terms of realism and input coherence in generating images with multimodal conditioning. Source code and trained models are available at: \url{https://github.com/aimagelab/Ti-MGD}.
\end{itemize}

%% file: sections/02_related.tex
\tinytit{Text-Guided Image Generation} 
The objective of text-to-image synthesis is to create an image that accurately represents a given textual prompt. Initially, approaches in this domain relied on GANs~\cite{xu2018attngan,zhu2019dm,zhang2021cross,tao2022df}, while recent advancements have shifted towards the use of diffusion models~\cite{nichol2022glide,ramesh2022hierarchical,rombach2022high,podell2023sdxl}. Among them, Nichol~\etal~\cite{nichol2022glide} proposed a text-to-image diffusion model with local editing capabilities to match more complex prompts. On a similar line, while Ramesh~\etal~\cite{ramesh2022hierarchical} proposed a two-stage approach involving a prior which produces CLIP image embeddings~\cite{Radford2021LearningTV} conditioned on textual captions and a diffusion-based decoder that translates images based on such embeddings, the approach proposed in~\cite{sahariaphotorealistic} leverages the T5 language model~\cite{raffel2020exploring} followed by a cascade of super-resolution diffusion models to improve the generation process. Unlike directly applying the diffusion process to pixel space, the current trend favors the use of latent diffusion models as introduced in~\cite{rombach2022high}. In these models, the forward and reverse processes are defined in the latent space of a pre-trained autoencoder, aiming for improved computational efficiency and the production of high-quality images.

Only a few attempts of text-to-image synthesis have been conducted for the fashion domain~\cite{zhu2017your,jiang2022text2human,pernuvs2023fice}. Notably, Zhu~\etal~\cite{zhu2017your} introduced a GAN-based solution that generates the final image based on both textual data and semantic layouts. A different approach is the one presented in~\cite{pernuvs2023fice}, where a latent code regularization technique is employed to enhance the GAN inversion process. This involves leveraging CLIP textual embeddings~\cite{Radford2021LearningTV} to guide the image generation process. Differently, Jiang~\etal~\cite{jiang2022text2human} proposed to synthesize full-body images by mapping textual descriptions of clothing items into one-hot vectors. However, this approach imposes limitations on the expressive capacity of the conditioning signal.

\tit{Multimodal Image Generation with Diffusion Models}
A correlated set of studies seeks to incorporate various modalities into existing diffusion models, thereby enhancing control over the generation process~\cite{choi2021ilvr,meng2022sdedit,wang2022pretraining,mou2023t2i,cheng2023adaptively}. In this context, Choi~\etal~\cite{choi2021ilvr} proposed refining the generative mechanism of an unconditional denoising diffusion probabilistic model~\cite{nichol2021improved} by aligning each latent variable with a given reference image. Conversely, the approach proposed by Mang~\etal~\cite{meng2022sdedit} introduces noise to a stroke-based input and applies the reverse stochastic differential equation to generate images, without additional training. Instead, Wang~\etal~\cite{wang2022pretraining} suggested learning a deeply semantic latent space and conducting conditional fine-tuning for each downstream task to correlate guidance signals with the pre-trained space. Other recent studies have suggested incorporating sketches as additional conditioning signals, either by concatenating them with the model input~\cite{cheng2023adaptively} or by training an MLP-based edge predictor to map latent features to spatial maps~\cite{voynov2022sketch}.

Among contemporary works designing novel conditioning strategies for pre-trained latent diffusion models, Zhang~\etal~introduced ControlNet~\cite{zhang2023adding} which extends Stable Diffusion~\cite{rombach2022high} with an additional conditioning input.
This procedure entails creating two variants of the original model parameters: one remains fixed and unaltered (referred to as the locked copy), while the other can be updated during training (referred to as the trainable copy).
The objective is to enable the trainable version to assimilate the newly introduced condition, while the locked version preserves the knowledge of the original model.
Conversely, the model proposed in~\cite{mou2023t2i} incorporates modality-specific adapter modules facilitating Stable Diffusion conditioning on novel modalities. On a similar line, Ye~\etal~\cite{ye2023ip} introduces a lightweight adapter that can enable Stable Diffusion conditioning on image prompts, directly using the cross-attention layers of the denoising network. In contrast, our focus lies within the fashion domain, where we propose a human-centric architecture based on latent diffusion models, leveraging direct conditioning from textual sentences and other modalities like human body poses, garment sketches, and fabric textures. 

\tit{Textual Inversion}
Textual inversion, as introduced in the recent work by Gal~\etal~\cite{gal2022textual}, is a novel technique aimed at learning pseudo words within the embedding space of a text encoder to represent visual concepts effectively. Building on this, several promising methods have been developed for personalized image generation and editing~\cite{han2023highly,ruiz2022dreambooth,daras2022multires,mokady2022null}. Among them, Ruiz~\etal~\cite{ruiz2022dreambooth} specifically introduced a fine-tuning technique that associates an identifier with a subject represented by a few images, incorporating a class-specific prior preservation loss to address language drift. Similarly, Kumari~\etal~\cite{kumari2022customdiffusion} proposed an alternative fine-tuning method for enabling multi-concept composition, demonstrating that updating only a small subset of model weights suffices to integrate new concepts. Instead, Han~\etal~\cite{han2023highly} decomposed the CLIP~\cite{Radford2021LearningTV} embedding space based on semantics, facilitating image manipulation without the need for further fine-tuning. In this work, we adapt textual inversion techniques to effectively condition latent diffusion models on garment fabric textures.

\tit{Virtual Try-On}
Another related research area involves the virtual try-on of a desired garment~\cite{han2018viton}. This task is usually addressed by first generating the warped version of the input garment and then synthesizing the final image of a reference person wearing the warped clothing item while preserving human pose and identity. Research efforts in this domain have mainly focused on the improvement of the geometric transformation phase~\cite{wang2018toward,yang2020towards,fincato2021viton} and on the design of additional components to enhance the realism of generated images~\cite{issenhuth2020not,ge2021parser,fele2022c}, also considering the problem in high-resolution settings~\cite{morelli2022dresscode,choi2021viton}. Recent attempts follow the latest trends in image generation and apply diffusion models to improve the quality of virtual try-on generated images~\cite{morelli2023ladi,li2023warpdiffusion,gou2023taming,zeng2023cat,kim2023stableviton}. While these approaches can be applied to enhance user experience, they all take the try-on garment as input and, therefore, can not directly be compared with the model proposed in this work, which instead generates the final image leveraging text, human pose, fabric texture, and sketch modalities.

%% file: sections/03_method.tex
\begin{figure*}[ht!]
\begin{center}
\includegraphics[width=\linewidth]{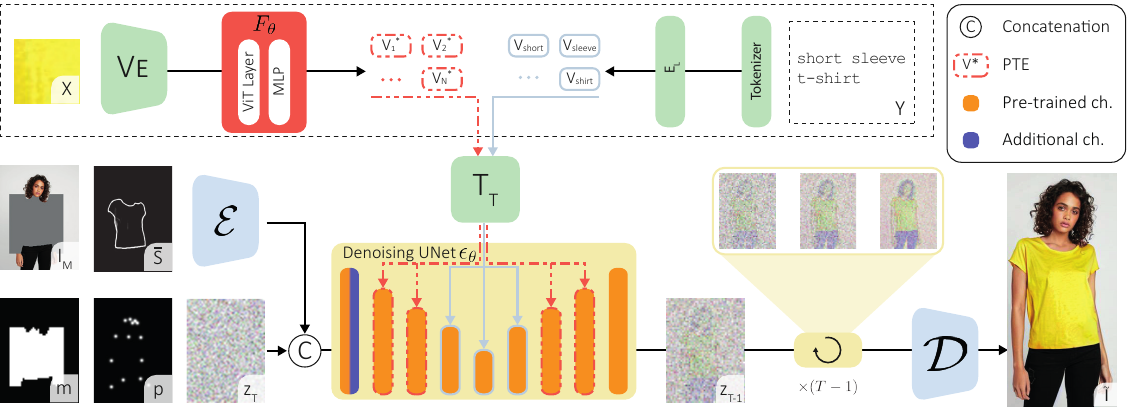}
\end{center}
\vspace{-0.1cm}
\caption{Overview of the proposed Textual-inverted Multimodal Garment Designer (\ours) approach, a human-centric latent diffusion model conditioned on multiple modalities, including text, human pose, garment sketch, and fabric texture. The denoising UNet $\epsilon_{\theta}$ takes as input the latent variable $z_T$ and the spatial conditioning inputs (\ie~encoded masked model $\encoder(\maskedimg)$, inpainting mask $\latentmask$, body keypoints $\latentpose$, and encoded sketch $\encoder(\bar{\sketch})$).
We incorporate text conditioning $\textual$ using Stable Diffusion cross-attention capabilities, extending this mechanism to condition the generated image on the texture image $X$ by projecting it into the CLIP pseudo-word token embedding space. For this purpose, we utilize distinct cross-attention layers dedicated to text and texture conditioning.}
\label{fig:model}
\vspace{-0.3cm}
\end{figure*}

This section proposes a novel task to automatically edit a human-centric fashion image conditioned on multiple modalities.
Specifically, given the model image $\img \in \mathbb{R}^{H \times W \times 3}$, its pose map $\pose \in \mathbb{R}^{H \times W \times 18}$ where each channel represent a human keypoint, a textual description $\textual$ of a garment, a sketch of the same $\sketch \in \mathbb{R}^{H \times W \times 1}$, and a sample image of a fabric texture $\texture \in \mathbb{R}^{H_X \times W_X \times 3}$, we want to generate a new image $\newimg \in \mathbb{R}^{H \times W \times 3}$ that retains the information of the input model while substituting the target garment according to the multimodal inputs.
To tackle the task, we propose a novel latent diffusion approach, denoted as Textual-inverted Multimodal Garment Designer (\ours), that effectively combines multimodal information when generating the new image $\newimg$.

To the best of our knowledge, this is the first proposed approach in literature to constrain fashion image editing on text, pose, sketch, and fabric texture.
We strongly believe this task can foster research in the field and enhance the design process of new fashion items with greater customization. An overview of our model is shown in Fig.~\ref{fig:model}.

\subsection{Preliminaries}
\tinytit{Stable Diffusion} While diffusion models~\cite{sohl2015deep} are latent variable architectures that work in the same dimensionality of the data (\ie~in the pixel space), latent diffusion models (LDMs)~\cite{rombach2022high} operate in the latent space of a pre-trained autoencoder achieving higher computational efficiency while preserving the generation quality. In our work, we leverage the Stable Diffusion model~\cite{rombach2022high}, a text-to-image implementation of LDMs, as a starting point to perform multimodal conditioning for human-centric fashion image editing. Stable Diffusion is composed of an autoencoder with an encoder $\encoder$ and a decoder $\decoder$, a text-time-conditional UNet denoising model $\epsilon_{\theta}$, and a CLIP-based text encoder $\CLIPtextenc$ taking as input a text $\textual$.
The encoder $\encoder$ compresses an image $\img$ into a lower-dimensional latent space defined in $\mathbb{R}^{h \times w \times 4}$, where $h=H/8$ and $w=W/8$. The decoder $\decoder$ performs the opposite operation, decoding a latent variable into the pixel space. For the sake of clarity, we define the $\epsilon_{\theta}$ convolutional input (\ie~$z_t$ in this case) as spatial input $\gamma$, because of the property of convolutions to preserve the spatial structure, and the attention conditioning input as $\psi$.
The denoising network $\epsilon_{\theta}$ is trained according to the following loss:
\begin{equation}
    L = \mathbb{E}_{\encoder(\img), \textual, \epsilon \sim \mathcal{N}(0,1),t} \left[ \lVert \epsilon - \epsilon_{\theta}(\gamma,\psi) \rVert_2^2 \right],
    \label{eq:diffusion_loss}
\end{equation}
where $t$ is the diffusing time step, $\gamma = z_t$, $\psi=\left[t;\CLIPtextenc(\textual)\right]$, and $\epsilon \sim \mathcal{N}(0,1)$ is the Gaussian noise added to $\encoder(\img)$.

\tit{CLIP} This vision-language model~\cite{Radford2021LearningTV} aligns visual and textual inputs in a shared embedding space. In particular, CLIP consists of a visual encoder $\CLIPvisenc$ and a text encoder $\CLIPtextenc = \CLIPlookup \circ \CLIPtexttrans$, where $\CLIPlookup$ is the embedding lookup layer, which maps each tokenized word of $\textual$ to the token embedding space $\mathcal{W}$, and $\CLIPtexttrans$ is the CLIP text Transformer that maps the token embedding features to the CLIP shared embedding space.
CLIP extracts feature representations $\CLIPvisenc(\img) \in \mathbb{R}^{d}$ and $\CLIPtextenc(\textual) \in \mathbb{R}^{d}$ for an input image $\img$ and its corresponding text caption $\textual$, respectively. Here, $d$ is the size of the CLIP shared embedding space. 

The proposed approach introduces a novel textual inversion technique to generate a representation of the fabric texture $\texture$. We feed this representation to the CLIP text Transformer $\CLIPtexttrans$ to condition the diffusion process. It consists in mapping the visual features of $\texture$ into a set of $N$ new token embeddings $V_n^* \in \mathcal{W}, n=\{1,\ldots, N\}$. 
Following the terminology introduced in~\cite{baldrati2023zeroshot}, we refer to these embeddings as Pseudo-word Tokens Embeddings (PTEs) since they do not correspond to any linguistically meaningful entity but rather are a representation of the fabric texture visual features in the token embedding space $\mathcal{W}$.

\subsection{Human-Centric Image Editing}
The proposed task aims to generate a new image $\newimg$, by replacing the target garment in the input image $\img$ using multimodal inputs while preserving the model's identity and physical characteristics. As a natural consequence, this task can be identified as a particular type of conditional inpainting tailored for human body data. 
Instead of using a standard text-to-image model, we perform inpainting concatenating along the channel dimension of the denoising network input $z_t$ an encoded masked image $\encoder(\maskedimg)$ and the relative resized binary inpainting mask $\latentmask \in \{0,1\}^{h \times w \times 1}$, which stems from the original inpainting mask $\mask \in \{0,1\}^{H \times W \times 1}$. Since here, the spatial input of the denoising network is $\gamma = [z_t; \latentmask; \encoder(\maskedimg)], \gamma \in \mathbb{R}^{h \times w \times 9}$.

To give users more precise control over the generation of garments, we propose extending the input capabilities of the denoising UNet by enabling constraints on multiple modalities. Essentially, we exploit spatial information such as pose and sketch to feed into the UNet spatial input $\gamma$, while we inject semantic information such as textual descriptions and fabric textures as attention conditioning input $\psi$. This allows for a more refined and accurate garment generation process.

The fully convolutional nature of the encoder $\encoder$ and the decoder $\decoder$ allows LDM-based architectures to preserve the spatial information in the latent space. Our method can thus optionally add conditioning constraints to the generation by exploiting this feature. In particular, we propose to add two spatial generation constraints: the model pose map $\pose$ to preserve the original human pose of the input model and the garment sketch $\sketch$ to condition the shape of the generated garment.

In addition, we leverage the Stable Diffusion textual information conditioning mechanism for two purposes: condition on plain text and condition on fabric texture information. While the former is intrinsic in the Stable Diffusion model by design, we propose a novel forward-only textual inversion method to tackle the latter without adding additional parameters in the denoising network.

\tit{Pose Map Conditioning}
In most cases~\cite{suvorov2022resolution,lugmayr2022repaint,li2022mat}, inpainting is performed with the objective of either removing or entirely replacing the content of the masked region. However, in our task, we aim to remove all information regarding the garment worn by the model while preserving the model's body information and identity.
Thus, we propose to improve the garment inpainting process by using the bounding box of the segmentation mask along with pose map information representing human body keypoints. This approach enables the preservation of the model's physical characteristics in the masked region while allowing the inpainting of garments with different shapes.

Differently from conventional inpainting techniques, we focus on selectively retaining and discarding specific information within the masked region to achieve the desired outcome.
To enhance the performance of the denoising network with human body keypoints,
we modify the first convolution layer of the network by adding 18 additional channels, one for each keypoint.
Adding new inputs usually would require retraining the model from scratch, thus consuming time, data, and resources, especially in the case of data-hungry models like the diffusion ones. 
Therefore, we propose extending the kernels of the pre-trained input layer of the denoising network and retraining the whole network. This consistently reduces the number of training steps, allowing training with less data. We extend these kernels using zero-initialized weights~\cite{zhang2023adding}, which allows us to retain the knowledge embedded in the original denoising network while enabling the model to deal with the newly proposed inputs.
Our experiments show that such improvement enhances the consistency of the body information between the generated in-painted region and the original image.

\tit{Incorporating Sketches}
Fully describing a garment using only textual descriptions is a challenging task due to the complexity and ambiguity of natural language. While text can convey specific attributes of a garment, like style and color, it may not provide sufficient information about its spatial characteristics, such as shape and size. This limitation can hinder the customization of the generated clothing item other than the ability to match the user's intended style accurately. Therefore, we propose to leverage garment sketches to enrich the textual input with additional spatial fine-grained details.
We achieve this following the same approach described for pose map conditioning.
The final spatial input of our denoising network is $\gamma = \left[z_t; \latentmask; \encoder(\maskedimg); \latentpose;\latentsketch\right]$, $\left[\latentpose; \latentsketch\right] \in \mathbb{R}^{h \times w \times (18+4)}$, where $\latentpose$ is obtained by resizing $\pose$ to match the latent space dimensions, while $\latentsketch = \encoder(\bar{\sketch})$ in which $\bar{\sketch}$ is the sketch $\sketch$ repeated along the channel dimension to match the $\encoder$ input channel shape. In the case of sketches, we only condition the early steps of the denoising process as the final steps have little influence on the shapes~\cite{balaji2022ediffi}.

\tit{Adding Texture}
While text conditioning can provide a high-level constraint over the generated garment style, it still misses the ability to express the high-frequency visual details of the garment fabric. This requirement is fundamental to give the user fine-grained control over the garment generation. We propose to enable the model to generate a garment coherent with a user-given fabric texture sample, denoted as $\texture$.

Starting with a given fabric texture sample image $\texture$, our objective is to condition the generation of the LDM utilizing the non-constrained receptive field of the attention mechanism. As the texture sample lacks spatial information and is intended to serve as a pattern for the generated garments, we propose using the existing cross-attention layers originally trained for textual conditioning, thus avoiding additional layers in the denoising network.
To this aim, starting from a given fabric texture sample image $\texture$, we leverage a forward-only textual inversion technique to predict a set of fine-grained Pseudo-word Token Embeddings (PTEs) describing the fabric texture $\texture$ itself. These PTEs are processed by the CLIP text transformer $\CLIPtexttrans$ to generate feature vectors that can condition the diffusion model generation.
In particular, we feed a given fabric texture sample image $\texture$ to the CLIP visual encoder $\CLIPvisenc$ and extract the features of its last hidden layer. We learn to project these features into the CLIP token embedding space $\mathcal{W}$ as a set of PTEs $V^{*} = \{V_1^*, \ldots, V_N^*\}$. This is achieved by training a textual inversion adapter module $\tiadapter$. The overall mathematical formulation is as follows:
\begin{equation}
\label{eq:vstar}
V^{*} = \{V_1^*, \ldots, V_N^*\} = \tiadapter(\CLIPvisenc(\texture)).
\end{equation} 
We then use the predicted PTEs $V^{*}$ to condition the Stable Diffusion denoising network $\epsilon_{\theta}$ and obtain the final image $\newimg$ where the model in $I$ is wearing the garment filled with the texture $\texture$. For clarity, a set of PTEs represents a fabric texture well if the model conditioned on the predicted pseudo-words can reconstruct the fabric texture of the target image itself.

We leverage the intuition in~\cite{voynov2023p+} where distinct Stable Diffusion cross-attention layers capture diverse granularity conditioning details, introducing an innovative approach. Our method enables concurrent textual and texture generation conditioning by leveraging the inherent capabilities of the existing Stable Diffusion layers. Importantly, this strategy avoids the introduction of extra parameters, ensuring a streamlined and efficient process.
To the best of our knowledge, this study marks the first instance in which a textual inversion approach is used for texture conditioning in the fashion image generation domain. The proposed approach diverges from conventional textual inversion methods like~\cite{gal2022textual, ruiz2022dreambooth, kumari2022customdiffusion}. Instead of iteratively optimizing pseudo-word token embeddings, our solution trains the adapter $\tiadapter$ to generate these embeddings in a single forward pass.

\subsection{Training and Inference}
Following the standard LDM approach, the proposed denoising network predicts the noise added stochastically to the encoded input, $z = \mathcal{E}(I)$. The objective function can be specified as
\begin{equation}
\label{eq:our_loss}
    L = \mathbb{E}_{\mathcal{E}(I), Y, \epsilon \sim \mathcal{N}(0,1), t, \mathcal{E}(I_M),m,p,s, V^*} \left[ \lVert \epsilon - \epsilon_{\theta}(\gamma,\psi) \rVert_2^2 \right],
\end{equation}
where $\gamma=\left[z_t; m; \mathcal{E}(I_M); p; s\right]$ and $\psi=\left[t; T_E(Y); \CLIPtexttrans(V^*)\right]$.

\tit{Classifier-Free Guidance}
Classifier-free guidance is a technique in which the denoising network works both conditionally and unconditionally.
This procedure adjusts the final predicted noise of the model so that it moves from the predicted unconditional noise toward the direction of the predicted conditioned one. Given the time step $t$ and a generic condition $c$, the predicted diffusion process follows the below equation:
\begin{equation}
    \hat{\epsilon}_{\theta}(z_t | c) = \epsilon_{\theta}(z_t | \emptyset) + \alpha \cdot (\epsilon_{\theta}(z_t | c) - \epsilon_{\theta}(z_t | \emptyset)),
    \label{eq:diffusion_classifier_free}
\end{equation}
where $\epsilon_{\theta}(z_t | c)$ is the predicted noise at time $t$ given the condition $c$ and $\epsilon_{\theta}(z_t | \emptyset)$ is the predicted noise at time $t$ given the null condition. The guidance scale $\alpha$ is a hyperparameter that controls the degree of extrapolation towards the condition.

We use the fast variant multi-condition classifier-free guidance proposed in~\cite{avrahami2022spatext} to speed up the inference time while dealing with multiple input conditions (\ie~text, pose map, sketch, fabric texture).
Instead of performing the classifier-free guidance according to each condition independently, the fast classifier-free guidance computes the direction considering all the conditions jointly $\Delta_{\text{joint}}^t = \epsilon_{\theta}(z_t | \{ c_i \}_{i=1}^{i=K}) - \epsilon_{\theta}(z_t | \emptyset)$:
\begin{equation}
    \hat{\epsilon}_{\theta}(z_t | \{ c_i \}_{i=1}^{i=K}) = \epsilon_{\theta}(z_t | \emptyset) + \alpha \cdot \Delta_{\text{joint}}^t. 
    \label{eq:diffusion_classifier_free2}
\end{equation}
where $K$ is the number of conditioning prompts.
This reduces the number of feed-forward executions from $K+1$ to $2$.

\tit{Unconditional Training}
To enhance the performance of the denoising model with and without specific conditions, we randomly drop them at training time. This method enables the model to adapt to both conditional and unconditional samples, enhancing mode coverage and sample fidelity. Additionally, it allows for the optional use of conditioning signals at inference time. Since our approach considers several conditioning signals, we propose to mask each condition independently. Our experiments demonstrate that adjusting the amount of masked data can significantly improve the output quality.

\begin{figure}[t]
    \centering
    \includegraphics[width=\linewidth]{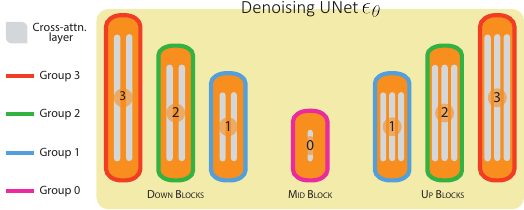}
    \caption{Detail of cross-attention layers of the denoising network, that are categorized into four groups based on spatial resolution. Group 3 contains the highest-resolution layers, while Group 0 comprises the lowest-resolution ones.}
    \label{fig:cross_attention_groups}
    \vspace{-0.3cm}
\end{figure}

\tit{Inference Modality-Aware Group Conditioning} 
In our task, we want to condition the generation on multiple prompts. As already stated, we cluster the input prompts in two groups: inputs with spatial information $\gamma$, that we feed to the denoising network convolutional input, and attention conditioning input $\psi$, that contain only semantic information and that leverage the cross-attention conditioning. Since the fabric texture does not contain spatial information, we categorize it as an attention conditioning input $\psi$.
We leverage the existing cross-attention blocks originally trained for textual conditioning to avoid adding additional parameters in the denoising UNet and reduce the computational load (\ie~layers inside the denoising network $\epsilon_\theta$ are executed $T$ times during inference, where $T$ is the number of the denoising steps). 

Our idea builds upon the intuition that different cross-attention layers in the Stable Diffusion denoising UNet process the input prompts differently according to the layer resolution~\cite{voynov2023p+}. More in detail, higher resolution layers (\ie~external layer in the UNet architecture) capture small-level details, while lower-resolution layers (\ie~internal layers) capture more coarse information, like shapes.
Therefore, we propose to condition the generation using the fabric texture information in the higher-resolution layers and textual information in the lower-resolution ones. This allows the condition of the generation on both textual and texture prompts without losing input information. We experimentally show that leveraging the fabric texture conditioning in each cross-attention layer leads to comparable results to conditioning only external ones.
Specifically, given the denoising network $\epsilon_\theta$, we categorize its attention layers into four groups, as illustrated in Fig.~\ref{fig:cross_attention_groups}. We name Group 3 the cross-attention layers with the highest resolution (the outermost layers) and sequentially assign lower group numbers down to Group 0 as the resolution decreases. Group 0 comprises the lowest resolution cross-attention layers (the innermost layers).

To maintain the flexibility of independently conditioning the generation on each modality (\eg~exclusively on texture or text), we adopt a training strategy involving prompt-exclusive conditioning alternation across samples. In other words, each sample is trained using either exclusive text conditioning or texture conditioning across all cross-attention layers.

%% file: sections/04_dataset.tex
\begin{figure}[t]
\begin{center}
    \includegraphics[width=\linewidth]{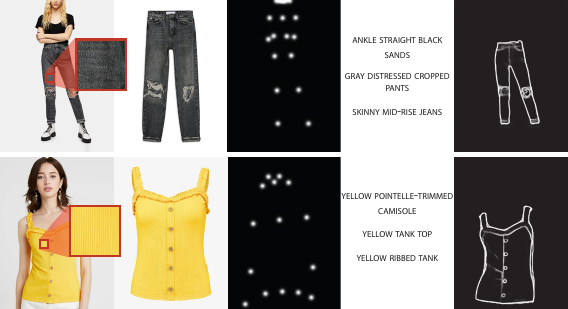}
\end{center}    
\vspace{-0.1cm}
\caption{Sample images and multimodal data from our newly collected \dataset and \datasetviton datasets.}
\label{fig:dataset}
\vspace{-.3cm}
\end{figure}

Current fashion image generation datasets often feature low-resolution images and lack the necessary multimodal information for the task we want to address. Therefore, creating new multimodal datasets is essential for advancing research in the fashion domain. To this aim, we start from two recent high-resolution fashion datasets, Dress Code~\cite{morelli2022dresscode} and VITON-HD~\cite{choi2021viton}, used for virtual try-on, and extend them by adding textual descriptions, garment sketches, and fabric textures. Both datasets contain image pairs with a resolution of $1024\times768$, each composed of a garment image and a corresponding model image wearing it. In this section, we present a framework for semi-automatically adding multimodal information to fashion images, detailing how we extend the Dress Code and VITON-HD datasets with multimodal annotations. The extended versions of these datasets are named \dataset and \datasetviton, respectively. Examples of images and multimodal data from these datasets are shown in Fig.~\ref{fig:dataset}.

\subsection{Dataset Collection and Annotation}

\tinytit{Data Preparation}
We begin annotating the Dress Code dataset, which comprises over 53k model-garment pairs across various categories. The initial task is to annotate each garment with a concise yet detailed textual description, using fashion-specific and non-generic terms suitable for guiding the generation process. Inspired by research indicating that people typically describe fashion items in just a few words~\cite{bianchi2021query2prod2vec}, we propose to use noun chunks. These are brief textual phrases consisting of a noun grouped together with its modifiers, effectively conveying key information while eliminating superfluous details and words.

Given the time-consuming and resource-intensive nature of manual annotation\footnote{Considering the Dress Code size of over 53k items and estimating 5 minutes per annotation, a single annotator working 8 hours a day, 5 days a week, for 260 days a year would need over 2 years to complete the task.}, we propose a semi-automatic framework using noun chunks for annotating the dataset. 
Initially, we gather domain-specific captions from two existing fashion datasets, FashionIQ~\cite{wu2021fashion} and Fashion200k~\cite{han2017automatic}. These captions are then standardized using word lemmatization, with each word reduced to its root form using the NLTK library\footnote{\href{https://www.nltk.org/}{https://www.nltk.org/}}. Subsequently, we extract noun chunks from these standardized captions, eliminating the article at the beginning of each noun chunk if present and any textual elements that contain or start with special characters. This pre-processing step yields over 60,000 unique noun chunks, categorized into three groups: upper-body clothes, lower-body clothes, and dresses. Table~\ref{tab:iqand200_numbers} reports detailed statistics about the number of unique captions and extracted noun chunks from which we start the annotation.

We then use the CLIP model~\cite{Radford2021LearningTV} and its open-source counterpart, OpenCLIP~\cite{wortsman2022robust}, to identify the most relevant noun chunks for each garment. For the CLIP model, we choose the ViT-L/14@336px and RN50$\times$64 versions and for OpenCLIP, we use the ViT-L/14, ViT-H/14, and ViT-g/14 models. To improve accuracy, we employ prompt ensembling. For each image, we associate 25 unique noun chunks by selecting the top-5 from each model based on the cosine similarity between the image and text embeddings, avoiding any duplicates.

\tit{Human-Labeled Textual Annotation}
To guarantee the accuracy and diversity of our annotations, we manually label a substantial portion of the Dress Code dataset images. Specifically, we select the three most fitting noun chunks from the 25 automatically associated ones for each garment image. We develop a custom tool to streamline the annotation process, limiting the average annotation time to 60 seconds per item. This tool also allows the manual insertion of noun chunks when the automatic choices are unsuitable. We manually annotate 26,400 garments (8,800 in each category) from the total 53,792 products in the dataset, ensuring the inclusion of all fashion items from the original test set~\cite{morelli2022dresscode}.

\tit{Hybrid Textual Annotation}
To finalize the textual annotation process, we first fine-tune the OpenCLIP ViT-B/32 model pre-trained on the English portion of the LAION-5B dataset~\cite{schuhmann2022laionb}, using our newly annotated image-text pairs. Next, we use this updated model and the extracted noun chunks to automatically tag the remaining items in the Dress Code dataset with the three most relevant noun chunks, chosen based on cosine similarity between multimodal embeddings. We apply the same method for automatically annotating all upper-body garment images in the VITON-HD dataset, restricting the noun chunks to those specifically describing upper-body clothing.

\input{tables/dataset/iqand200_numbers}

\tit{Extracting Sketches} 
Introducing garment sketches enhances the design details in our dataset, offering a more accurate and comprehensive view that text alone cannot express. This leads to better quality and control in the generated designs. We use PiDiNet~\cite{su2021pixel}, a pre-trained edge detection network, to extract sketches for both the Dress Code and VITON-HD datasets.

The datasets, originally introduced for virtual try-on, include paired and unpaired test sets. In the paired setting, each sample features an in-shop garment and a model wearing it, while in the unpaired setting, the in-shop garment is different from the one the model is wearing. For the paired set, we use the human parsing mask to isolate the garment on the model before processing it with the edge detection network. For the unpaired set, since we do not have a warped reference garment, we first match the in-shop garment to the model's pose and shape. To this aim, we employ a geometric transformation module that leverages a thin-plate spline transformation~\cite{rocco2017convolutional} and refines the results with a UNet model~\cite{ronneberger2015u}. This allows us to extract sketches from each warped garment, making our approach also applicable to unpaired settings.

In particular, we follow recent virtual try-on literature~\cite{issenhuth2020not,yang2020towards} and implement a warping module that computes a correlation map using encoded representations of the in-shop garment $C$ and a cloth-agnostic person representation, combining the pose map $P \in \mathbb{R}^{H \times W \times 18}$ and the masked model image $I_M \in \mathbb{R}^{H \times W \times 3}$. These representations are obtained through two distinct convolutional networks. The correlation map then guides the prediction of spatial transformation parameters for a thin-plate spline geometric transformation. Using these parameters, we generate a coarse warped garment $\hat{C}$ from $C$. A UNet model further refines this output, taking as input the coarse warped garment $\hat{C}$, the pose map $P$, and the masked model image $I_M$, to produce the finely warped garment $\Tilde{C}$.

\tit{Extracting Textures}
To enable users precise control over garment generation, including the fabric texture samples in the dataset is crucial, as text conditioning cannot convey detailed visual fabric characteristics. 
Given an in-shop garment $C$ and its garment mask $M_C$, we extract fabric textures leveraging a sliding window mechanism. For each in-shop garment $C$ and its corresponding mask $M_C$, we extract fabric textures using a sliding window of $128\times128$ pixels, selecting only patches $X$ fully within the garment mask $M_C$. To prevent patch redundancy, we employ a stride of $\frac{128}{2} = 64$ pixel horizontally and vertically. We use high-resolution dataset images (\ie~$1024\times768$ pixel) for this process.
When the algorithm cannot find a suitable texture patch (\eg~mostly in short pants), we reduce the window size to $64\times64$ pixels to guarantee at least one patch $X$ for each garment $C$.

\input{tables/dataset/comparison.tex}

\subsection{Comparison with Other Datasets}
The only two text-to-image generation datasets in the fashion domain, referenced in~\cite{zhu2017your} and~\cite{jiang2022text2human}, both utilize images from the DeepFashion dataset~\cite{liu2016deepfashion}. The dataset from~\cite{zhu2017your} includes brief textual descriptions, while DeepFashion-Multimodal~\cite{jiang2022text2human} features attributes (\eg~category, color, fabric, etc.) for crafting longer captions. In Table~\ref{tab:datasets_comparison}, we compare the textual annotation statistics of these publicly available datasets with our newly extended datasets. Along with the number of images and fashion products, we report the number of unique textual items, either noun chunks or textual sentences and the number of unique words excluding stop words and punctuation. Notably, our datasets exhibit a greater diversity in textual items and words, validating the effectiveness of our annotation approach and facilitating more customized control over the generation process. It is also important to note that the other datasets lack in-shop garment images, which limits their utility in our setting making it impossible to extract garment sketches for an unpaired and more realistic setting.

%% file: tables/dataset/iqand200_numbers.tex
\begin{table}[t]
\begin{center}
\caption{Number of unique captions and noun chunks for each category of the FashionIQ and Fashion200k datasets.}
\label{tab:iqand200_numbers}
\footnotesize
\setlength{\tabcolsep}{.32em}
\resizebox{\linewidth}{!}{
\begin{tabular}{lc ccc c ccc}
\toprule
& & \multicolumn{3}{c}{\textbf{Unique Captions}} & & \multicolumn{3}{c}{\textbf{Unique Noun Chunks}} \\
\cmidrule{3-5} \cmidrule{7-9}
\textbf{Dataset} & & \textbf{Upper} & \textbf{Lower} & \textbf{Dresses} & & \textbf{Upper} & \textbf{Lower} & \textbf{Dresses} \\
\midrule
FashionIQ~\cite{wu2021fashion} & & 27,339 & 0 & 15,101 & & 7,801 & 0 & 3,592  \\
Fashion200k~\cite{han2017automatic} & & 25,959 & 11,022 & 16,694 & & 22,898 & 13,420 & 15,890 \\
\bottomrule
\end{tabular}
}
\end{center}
\vspace{-0.4cm}
\end{table}

%% file: tables/dataset/comparison.tex
\begin{table}[t]
\begin{center}
\caption{Comparison of \dataset and \datasetviton with other fashion datasets featuring multimodal annotations. Here T stands for Text, P for Pose, S for Sketch, and F for Fabric texture.}
\label{tab:datasets_comparison}
\footnotesize
\setlength{\tabcolsep}{.3em}
\resizebox{\linewidth}{!}{
\begin{tabular}{lc cc c cc cc}
\toprule
& & & & & & & \textbf{\# Unique} & \textbf{\# Unique} \\
\textbf{Dataset} & \textbf{T} & \textbf{P} & \textbf{S} &  \textbf{F} & \textbf{\# Images} & \textbf{\# Products} & \textbf{Texts} & \textbf{Words} \\
\midrule
VITON-HD~\cite{choi2021viton} & \xmark & \cmark & \xmark & \xmark & 27,358 & 13,679 & - & - \\
Dress Code~\cite{morelli2022dresscode} & \xmark & \cmark & \xmark & \xmark & 107,584 & 53,792 & - & - \\
\midrule
Be Your Own Prada~\cite{zhu2017your} & \cmark & \cmark & \xmark & \xmark & 78,979 & N/A & 3,972 & 445 \\
DF-Multimodal~\cite{jiang2022text2human} & \cmark & \cmark & \xmark & \xmark & 44,096 & N/A & 10,253 & 77 \\
\midrule
\rowcolor{blond}
\textbf{\datasetviton} & \cmark & \cmark & \cmark & \cmark & 27,358 & 13,679 & 5,143 & 1,613 \\
\rowcolor{blond}
\textbf{\dataset} & \cmark & \cmark & \cmark & \cmark & 107,584 & 53,792 & 25,596 & 2,995 \\
\bottomrule
\end{tabular}
}
\end{center}
\vspace{-0.35cm}
\end{table}

%% file: sections/05_experiments.tex
\subsection{Implementation Details and Competitors}
\tinytit{Training and Inference}
All models are trained on the original splits of the Dress Code~\cite{morelli2022dresscode} and VITON-HD~\cite{choi2021viton} datasets using a single NVIDIA A100 GPU. Specifically, Dress Code contains around 48k training items and 5,400 test ones, instead VITON-HD is divided into 11,647 and 2,032 products respectively belonging to the training and test set.

In all experiments, we use an image resolution of $512 \times 384$. When trained on \dataset, models undergo 200k training steps, while for \datasetviton, they are trained for 75k steps. 
As the latent diffusion model, we use Stable Diffusion inpainting v2\footnote{\href{https://huggingface.co/stabilityai/stable-diffusion-2-inpainting}{https://huggingface.co/stabilityai/stable-diffusion-2-inpainting}}. To ensure a fair comparison with other models, we also develop a version of \ours based on Stable Diffusion inpainting v1\footnote{\label{sd1}\href{https://huggingface.co/runwayml/stable-diffusion-inpainting}{https://huggingface.co/runwayml/stable-diffusion-inpainting}}.
During training, we use a batch size of 16 and a learning rate of $10^{-5}$, with a linear warm-up in the first 500 iterations. AdamW~\cite{loshchilov2019decoupled} is employed as optimizer, with a weight decay of $10^{-2}$. To speed up training and save memory, we use mixed precision~\cite{micikevicius2018mixed}. 
We set the unconditional portion of data during training and the sketch conditioning rate during inference to $0.2$ each.

Training involves using textual conditions half of the time and texture conditioning for the remaining half, allowing the network to adapt to both independently. At inference time, when we leverage both texture and textual conditions, we use textual features to condition Groups 0 and 1, while Groups 2 and 3 are conditioned with texture features, following the notation of Fig.~\ref{fig:cross_attention_groups}.
The textual inversion network $F_{\theta}$ comprises a single ViT layer and an MLP projection. The MLP includes three fully connected layers, each separated by GELU non-linearity~\cite{hendrycks2016gaussian} and a dropout layer~\cite{srivastava2014dropout}. The network outputs $N=16$ PTEs. As the visual encoder $V_{E}$, we use OpenCLIP ViT-H/14~\cite{wortsman2022robust}, pre-trained on the English portion of the LAION-5B dataset~\cite{schuhmann2022laionb}.
We use the DDIM~\cite{song2021denoising} with 50 steps as our noise scheduler during inference, setting the classifier-free guidance parameter $\alpha$ to 7.5. To improve the high-frequency details in the region outside the inpainting area, we leverage the EMASC module defined in~\cite{morelli2023ladi}.

\tit{Baselines and Competitors} 
To ensure fair comparisons between our model and competitors, we train a version of our model using the same backbone of the competing approaches and compare results against approaches specialized on different subsets of modalities.
For text-only inputs, we compare \ours with the Stable Diffusion inpainting pipeline available on Huggingface\footref{sd1}. In scenarios involving text and pose inputs, \ours is compared with Stable Diffusion v1.5 integrated with ControlNet~\cite{zhang2023adding} for pose conditioning\footnote{\href{https://huggingface.co/lllyasviel/control\_v11p\_sd15\_openpose}{https://huggingface.co/lllyasviel/control\_v11p\_sd15\_openpose}}. For inputs of text, pose, and sketch, we set \ours against an adapted version of SDEdit~\cite{meng2022sdedit} and Stable Diffusion v1.5 integrated with ControlNet with pose and sketch adapters\footnote{\href{https://huggingface.co/lllyasviel/control\_v11p\_sd15\_softedge}{https://huggingface.co/lllyasviel/control\_v11p\_sd15\_softedge}}. Specifically for SDEdit, we follow the approach in~\cite{meng2022sdedit}, using our model trained with only text and human poses and guiding the shape with a noise-added sketch image as the starting latent variable, setting the strength parameter to $0.9$.
For completeness, we also include the results of the previous version of our model, \ie~MGD~\cite{baldrati2023multimodal}.
For the full input set modalities (\ie~text, pose, sketch, and texture), we employ ControlNet for text, pose, and sketch, and the IP-Adapter~\cite{ye2023ip}\footnote{\href{https://huggingface.co/h94/IP-Adapter}{https://huggingface.co/h94/IP-Adapter}} for texture, as ControlNet handles only inputs with spatial information. We set the conditioning scale for all ControlNet networks at $0.5$ and condition on sketches for only the first $0.2$ fraction of denoising steps. The IP-Adapter scale is set to $0.8$. Note that our proposed methods and IP-Adapter both leverage OpenCLIP ViT-H/14 as the visual encoder.

\input{tables/model/main}

\subsection{Evaluation Metrics}
\label{sec:metrics}
To evaluate the realism of generated images, we use the Fréchet Inception Distance (FID)~\cite{heusel2017gans} and the Kernel Inception Distance (KID)~\cite{binkowski2018demystifying}, following the implementation proposed in~\cite{parmar2022aliased}. For assessing how well the images adhere to textual conditioning, we apply the CLIP Score (CLIP-S)~\cite{hessel2021clipscore} from the TorchMetrics library~\cite{detlefsen2022torchmetrics}, using the OpenCLIP ViT-H/14 model as cross-modal architecture. We compute the score on the inpainted region of the generated output pasted on a $224 \times 224$ white background. Additionally, we employ three evaluation metrics to assess the adherence of the generated image with respect to pose, sketch, and texture modalities.

\tit{Pose Distance (PD)} 
We introduce a novel pose distance metric to assess the consistency of human body poses between original and generated images by measuring the distance between the keypoints. 
Specifically, we employ the OpenPifPaf~\cite{kreiss2021openpifpaf} human pose estimation network and compute the $\ell_2$ distance between each pair of real-generated corresponding estimated keypoints. This metric focuses only on the keypoints within the generation mask $M$ and adjusts each keypoint distance based on the confidence scores from the detector to account for possible estimation inaccuracies.

\tit{Sketch Distance (SD)} 
To quantify the adherence to the sketch constraint, we propose a novel sketch distance metric. 
We first segment the generated garments using an off-the-shelf clothing segmentation network\footnote{\href{https://github.com/levindabhi/cloth-segmentation}{https://github.com/levindabhi/cloth-segmentation}}. Then, we paste the segmented garment area onto a white background ($512\times384$) and use the PIDInet~\cite{su2021pixel} edge detector to extract sketches.
The final score is the mean squared error between the generated and input sketch $\sketch$, weighting each result by the inverse frequency of the activated pixels in $S$ to ensure a fair comparison. We avoid sketch thresholding to ensure a more effective comparison with hand-drawn grayscale sketches, enhancing the evaluation of sketch-guided image generation methods. 

\tit{Texture Score (TS)}
We introduce a new metric to assess how well the generated garment matches the input fabric texture. This similarity is determined by extracting and comparing visual features from the input patch and the generated garment texture. We use the same segmentation network as in the sketch distance calculation to isolate the garment information. Then, we crop a $64\times64$ portion of the image to represent the texture of the generated garment. The adherence of the generated image texture to the input is evaluated using the CLIP cosine similarity with the OpenCLIP ViT-H/14 model, the same model used for computing the CLIP score.

\subsection{Experimental Results}

\input{tables/model/main_categories}

\tinytit{Comparison with Other Methods}
We test our proposed method for each dataset under paired and unpaired settings. In the paired setting, conditions such as text, sketch, and fabric texture correspond to the garment worn by the model in the image. Conversely, in the unpaired setting, the input conditions relate to a different garment.
Table~\ref{tab:main_merged} presents quantitative results of our models benchmarked against the aforementioned competitors on \dataset and \datasetviton datasets. As it can be seen, the proposed \ours model consistently outperforms competitors in terms of realism (\ie~FID and KID) and coherency with input modalities (\ie~CLIP-S, PD, SD, and TS).

When considering text-only conditioned methods, we notice that Stable Diffusion~\cite{rombach2022high} can produce images fairly consistent with text conditioning, as underlined by the CLIP-S, while struggling to maintain the original model pose. Constraining the generation on pose using ControlNet~\cite{zhang2023adding} helps alleviate this issue, resulting in a lower pose distance while also boosting sketch distance and realism performances. We argue that the improvement related to SD depends on the correlation between the pose and the garment sketch, while the boost in realism stems from the additional details provided by the input. Incorporating sketch constraints shows mixed results when considering ControlNet~\cite{zhang2023adding} (row 4 in Table~\ref{tab:main_merged}) and SDEdit~\cite{meng2022sdedit}. The former slightly improves sketch coherence at the expense of realism, while SDEdit enhances both input coherence and realism. Note that we use our text-pose conditioned denoising network as the SDEdit backbone. When texture conditioning is added, we compare our \ours method against ControlNet combined with the IP-Adapter~\cite{ye2023ip}.
Notably, ControlNet+IP-Adapter not only boosts texture coherence metrics but also realism, as indicated by improved FID and KID scores. Nevertheless, \ours surpasses this combination in both realism and adherence to input conditions in both paired and unpaired settings. 
When comparing the proposed \ours approach using SDv1 and SDv2 as the backbone, we find comparable performance across all metrics, except in texture adherence, where the SDv2-based model shows a more significant improvement. Regarding instead the comparison with the previous version of our model, it is worth noting that adding texture conditioning leads to improved results across all metrics, except CLIP-S in which the previous version of our model achieves slightly improved performance. 

Table~\ref{tab:dresscode_categories} extends the previous analysis providing a detailed category-wise evaluation on the \dataset dataset. Due to the limited size of the test split for each category, containing only 1,800 images, the FID exhibits considerable variance in its results~\cite{binkowski2018demystifying}. In contrast, the KID delivers more reliable findings. Despite this, our method consistently surpasses all competitors across most metrics. The only exception is in the pose metrics under unpaired settings, which can be attributed to the challenges in aligning the predicted warped unpaired sketch with the model's body shape and pose. Also in this case, the previous version of our model achieves better results only in terms of CLIP-S.

\begin{figure*}[t]
\begin{center}
\includegraphics[width=\linewidth]{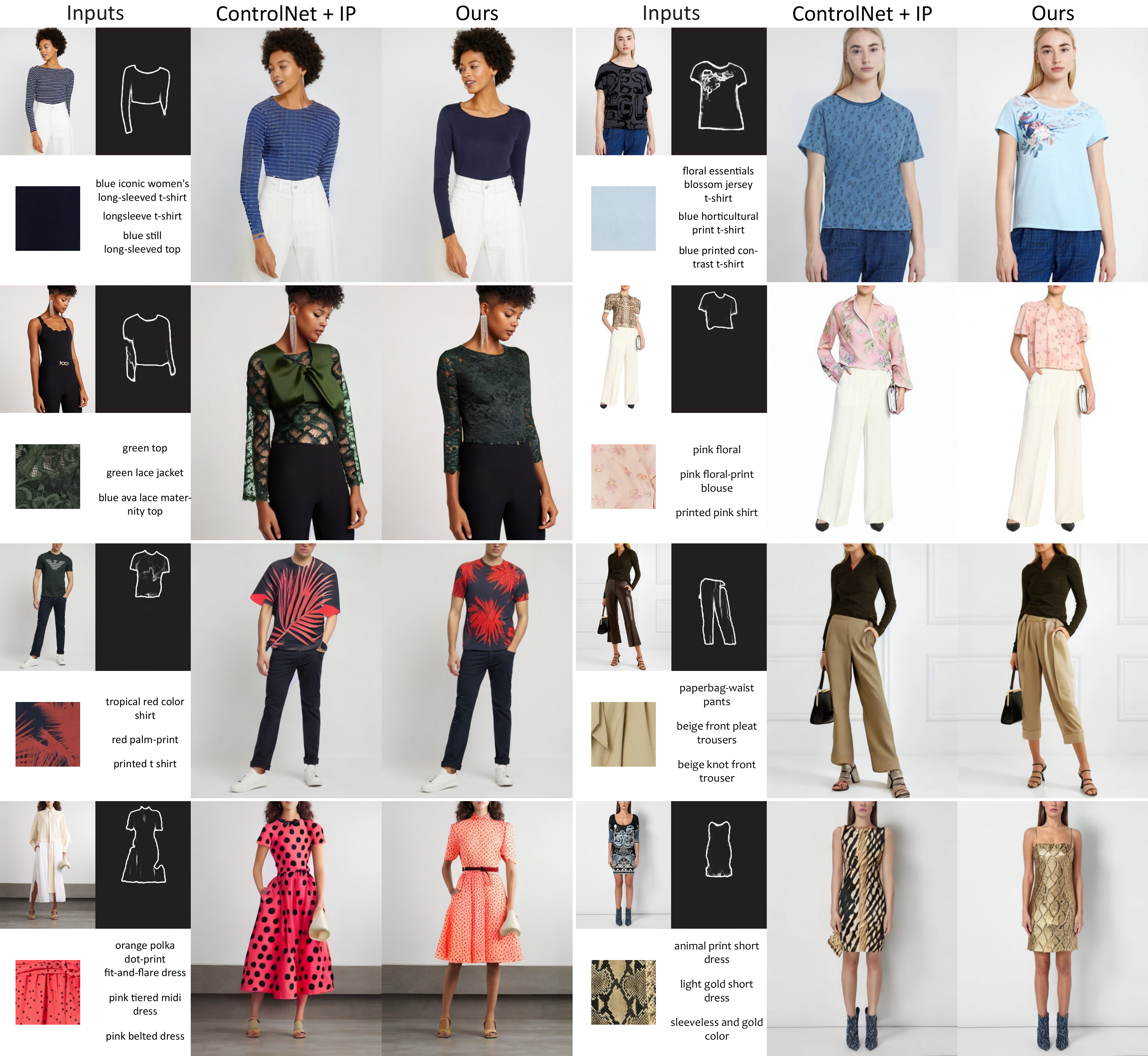}
\end{center}
\vspace{-0.2cm}
\caption{Qualitative comparison of images generated using our approach with SDv1 (\ours) versus ControlNet with IP-Adapter. The smaller images represent the model inputs, while the bigger images depict the generated outputs.} 
\label{fig:qualitative-methods}
\vspace{-0.35cm}
\end{figure*}

To qualitative validate our results, we show in Fig.~\ref{fig:qualitative-methods} a qualitative comparison between our proposed \ours and ControlNet+IP-Adapter. We report results generated by our method based on SDv1 for a fair comparison. Images in row1-col1 and row2-col2 show respectively the improved adherence of \ours model compared to the competitor respectively to the texture and sketch input. In row3-col1 and row4-col1, it is possible to see how the proposed method is able to combine text and texture information meaningfully, resulting in better input coherence than its competitor. Finally, images in row1-col2, row2-col1, row3-col2, and row4-col2 demonstrate the ability of \ours approach to blend the sketch and texture input, generating a realistic garment while following the multimodal input.

\input{tables/model/user_study}

\tit{User Study}
To validate our results with human feedback, we conducted a user study evaluating the realism and multimodal input adherence of generated images. Involving more than 100 participants, the study collected around 5,000 evaluations. Results displayed in Table~\ref{tab:user_study} reveal that our model consistently outperforms others in both realism and input adherence, confirming the effectiveness of our method. Comparing these results with previously reported quantitative evaluations highlights a correlation between our quantitative metrics and human judgments, both in terms of realism and coherence.

\tit{Varying Input Modalities}
In Table~\ref{tab:modalities_unpaired}, we analyze the behavior of our model with SDv2 as backbone when conditioned with various combinations of input modalities, either by masking inputs (\ie~using a zero tensor for pose and sketch) or by omitting them entirely (\ie~replacing texture cross-attention conditioning with text).
Notice that the text input anchors the CLIP-S metrics of all experiments and makes them comparable in all cases except when incorporating texture input. 
In this latter case, we observe a slight decrease in the CLIP-S due to the mismatch between the texture and textual information. 

\begin{figure*}[t]
\begin{center}
\includegraphics[width=\linewidth]{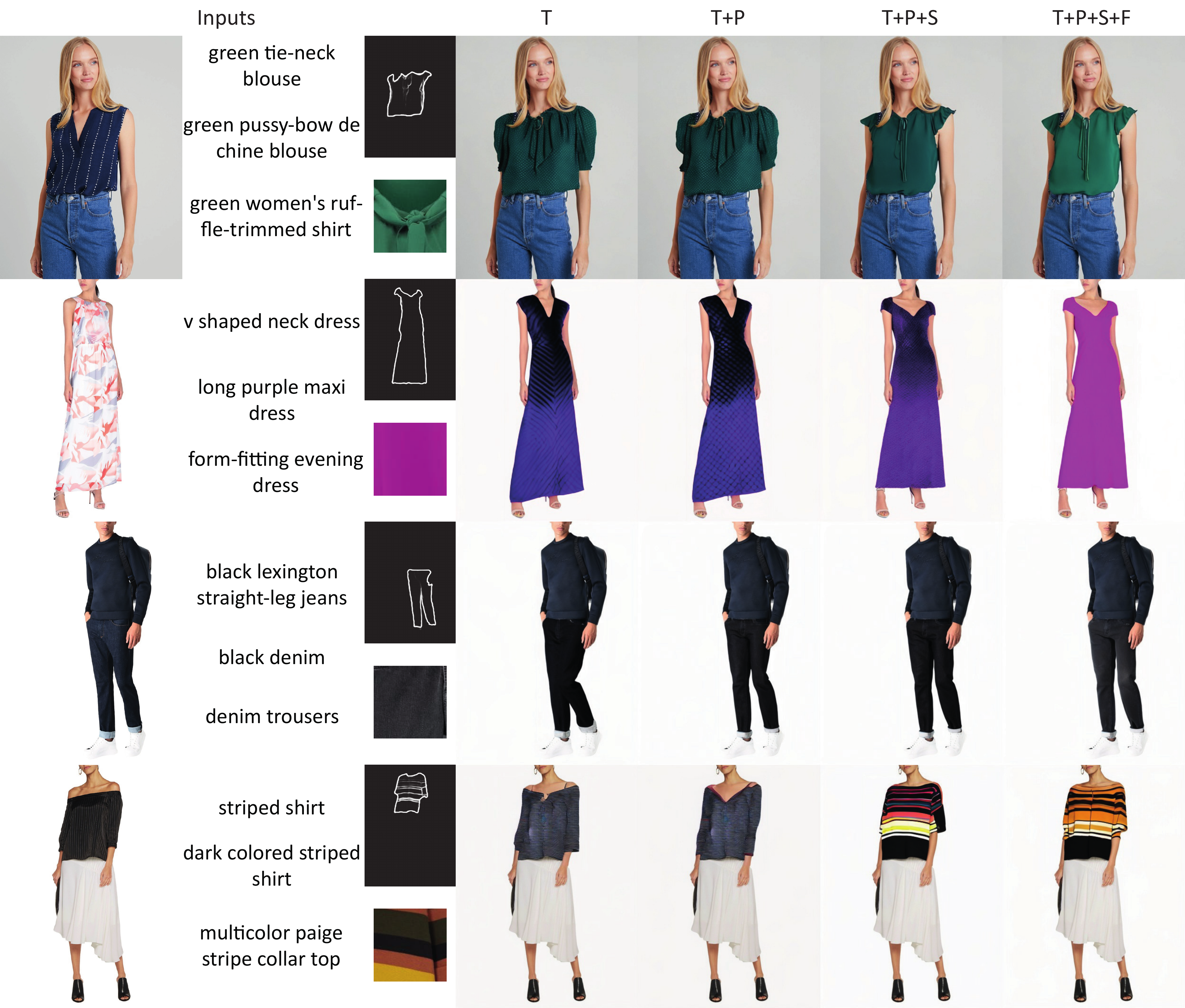}
\end{center}
\vspace{-0.2cm}
\caption{Qualitative examples of images generated by our proposed approach  (\ours) when varying the input modalities, where $T$ represents text, $P$ pose, $S$ sketch, and $F$ fabric texture.} 
\label{fig:qualitative-modalities}
\vspace{-0.35cm}
\end{figure*}

Starting from the fully conditioned model (\ie~text, pose, sketch, texture), we replace the texture conditioning with text. The decrease in the texture similarity confirms the impact of the texture input on the generation process in both \dataset and \datasetviton datasets. This also marginally affects the realism metrics (FID and KID), suggesting that texture information narrows the gap between generated and real images. Masking the sketch input leads to a higher sketch distance, underlining the capabilities of our model to incorporate this input effectively. In this case, the pose distance also increases slightly, reflecting the intertwined nature of the sketch and pose information, as the sketch implicitly includes details about the model's pose. Further masking the pose map input shows a decrease in the pose distance score, while realism metrics remain comparable. These results collectively demonstrate that adding modalities enhances the relative adherence metric, confirming the ability of our model to handle multiple conditions in a distinct manner efficiently. Fig.~\ref{fig:qualitative-modalities} shows from a qualitative point of view that masking the input modalities can affect the generated image.

\input{tables/model/modalities_ablation_unpaired}

\input{tables/model/uncond}

\input{tables/model/cross_attention}

\begin{figure*}[t]
\begin{center}
\includegraphics[width=\linewidth]{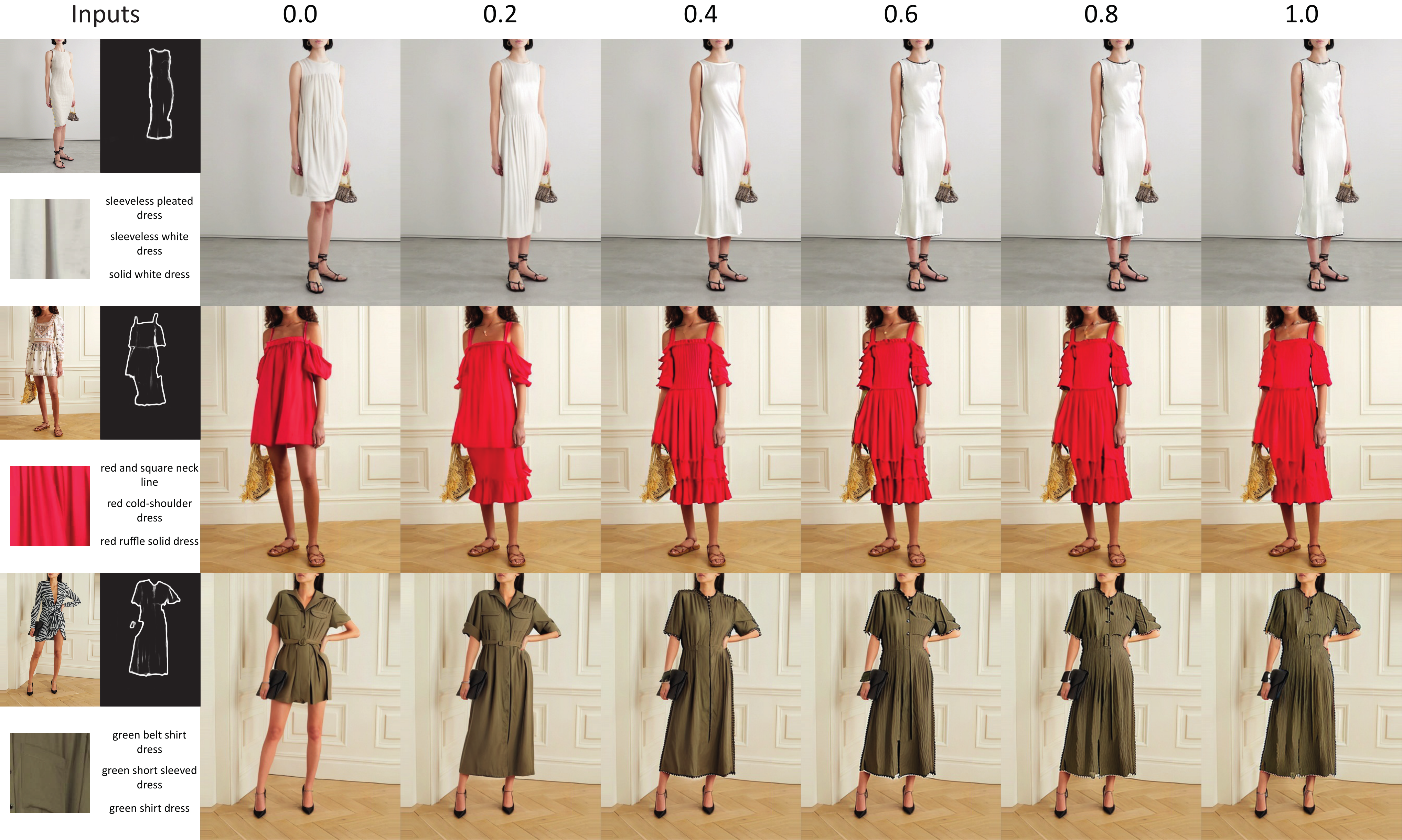}
\end{center}
\vspace{-0.2cm}
\caption{Qualitative examples of images generated by our proposed approach  (\ours) when varying the sketch conditioning rate. We report the such rate on top of each column. Results are reported on sample images from the \dataset dataset.} 
\label{fig:qualitative-sketch}
\vspace{-0.35cm}
\end{figure*}

\tit{Unconditional Training and Sketch Conditioning} Table~\ref{tab:dressCode_ablations} explores the performance of our fully conditioned network by varying the amount of unconditional training and the fraction of steps used to sketch conditioning. Specifically, we train three different models for unconditional training with fractions equal to $0.1$, $0.2$, and $0.3$. To evaluate the sketch conditioning rate, we test our model over a range from $0$ to $1$ with a stride of $0.2$. We find that optimal results are obtained when both parameters are set to $0.2$, providing an ideal balance between neglecting the sketch (at lower rates) and compromising realism (at higher rates). This is also confirmed from a qualitative point of view, as shown in Fig.~\ref{fig:qualitative-sketch}.

\tit{Inference Modality-Aware Group Conditioning}\label{sec:inference-conditioning}
Before analyzing the results, it is important to note that a textual description of a given texture image is a high-level representation of it. Hence, the same textual information can refer to hypothetically infinite texture images other than the given one, \eg~replicating a specific texture based on text alone can be challenging.
We can clearly see the effect of this asymmetry when analyzing images obtained conditioning only on text or texture.
For example, when creating images solely based on text input, the texture of the produced garment may not precisely replicate the specified texture image (Fig.~\ref{fig:qualitative-modalities}), resulting in a sub-optimal texture similarity score. However, this approach tends to maximize the CLIP-S. On the other hand, if the generation process focuses exclusively on the texture, the resulting images might closely resemble the intended texture, achieving a high texture similarity. Yet, this method might lead to a misalignment with the textual description, as evidenced by a lower CLIP-S. 

In other words, we argue that generating a garment that simultaneously maximizes CLIP-S and texture similarity is unfeasible since these metrics are correlated with the high-level semantic information while competing with the low-level visual details. In this scenario, we look for a sweet spot between the CLIP-S and texture similarity metrics.

\begin{figure*}[t]
\begin{center}
\includegraphics[width=\linewidth]{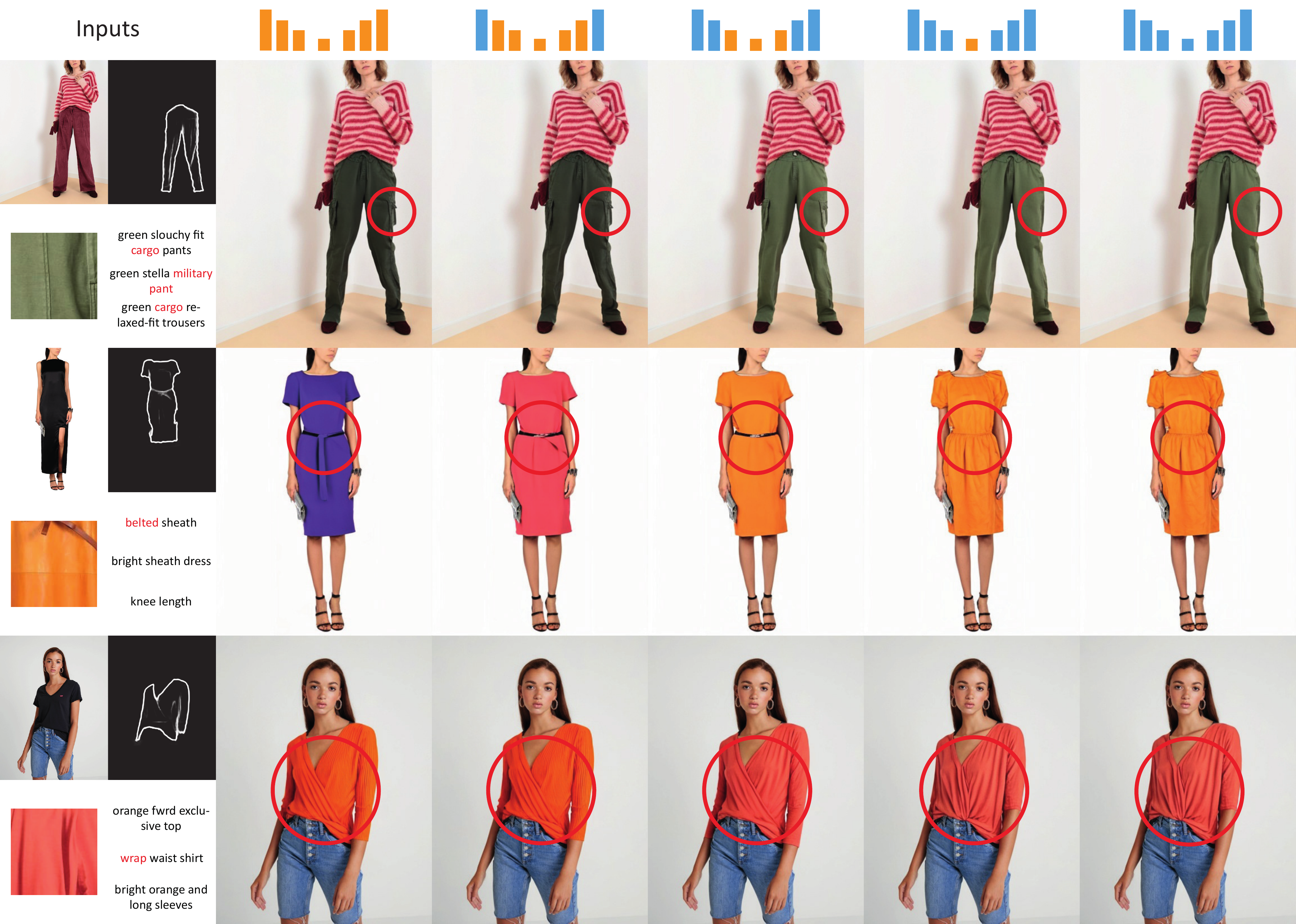}
\end{center}
\vspace{-0.2cm}
\caption{Qualitative examples of our proposed approach (\ours) on \dataset and \datasetviton datasets when using different cross-attention groups for conditioning on texture and text input. For a visual representation of the cross-attention groups see Fig.~\ref{fig:cross_attention_groups}.
At the top of the figure, we illustrate the conditioning modality of each cross-attention group.
Where, \tcbox[on line, boxsep=2pt, left=-1.2pt,right=-1.2pt,top=0pt,bottom=-0.5pt, sharp corners, colframe=white,colback=cross-attn-blue]{\textcolor{cross-attn-blue}{a}} stands for the cross-attention groups conditioned on the texture, while \tcbox[on line, boxsep=2pt, left=-1.2pt,right=-1.2pt,top=0pt,bottom=-0.5pt, sharp corners,
colframe=white,colback=cross-attn-yellow]{\textcolor{cross-attn-yellow}{a}} are cross-attention groups conditioned on the text. The red circles in the generated images highlight details derived from the textual input (also highlighted in red in the text). These details are visible when all cross-attention groups of the model are conditioned with textual information, as shown in the leftmost generated image in the figure. Conversely, they disappear in models fully conditioned on texture, as the textual information is no longer transmitted. On the contrary, the coherence of fabric texture is higher in the rightmost images and absent in the leftmost ones. Conditioning on Groups 3 and 2 (central column) emerges as a good trade-off between textual and texture fidelity, as evident from the images.} 
\label{fig:qualitative-cross-attn}
\vspace{-0.35cm}
\end{figure*}

We can observe this phenomenon in Table~\ref{tab:cross-annt-ablation}, which shows the performance of our network when varying the textual and texture conditioning across the cross-attention groups. It is worth noting that the CLIP-S and texture similarity scores depend on the number and position of cross-attention layer groups conditioned on text or texture, respectively. The more groups are conditioned on the texture, the higher the texture similarity, while the more groups are conditioned on the text, the higher the CLIP-S. However, if we consider experiments conditioned on the same number of groups on texture (\ie~rows 2 vs. 8, 3 vs. 7, or 4 vs. 6), we obtain higher texture similarities when we condition the outer groups on the texture image.
The best trade-off is obtained when the texture conditioning is applied to Groups 3 and 2 (\ie~row 3), which corresponds to a CLIP-S comparable to the only-text version and texture similarity comparable with the texture-only one. In Fig.~\ref{fig:qualitative-cross-attn}, we report some qualitative examples of images generated when performing texture conditioning across different cross-attention groups. It is possible to note that performing texture conditioning on Groups 3 and 2 allows the generation of an image in line with both textual and texture input information.

\begin{figure*}[t]
\begin{center}
\includegraphics[width=\linewidth]{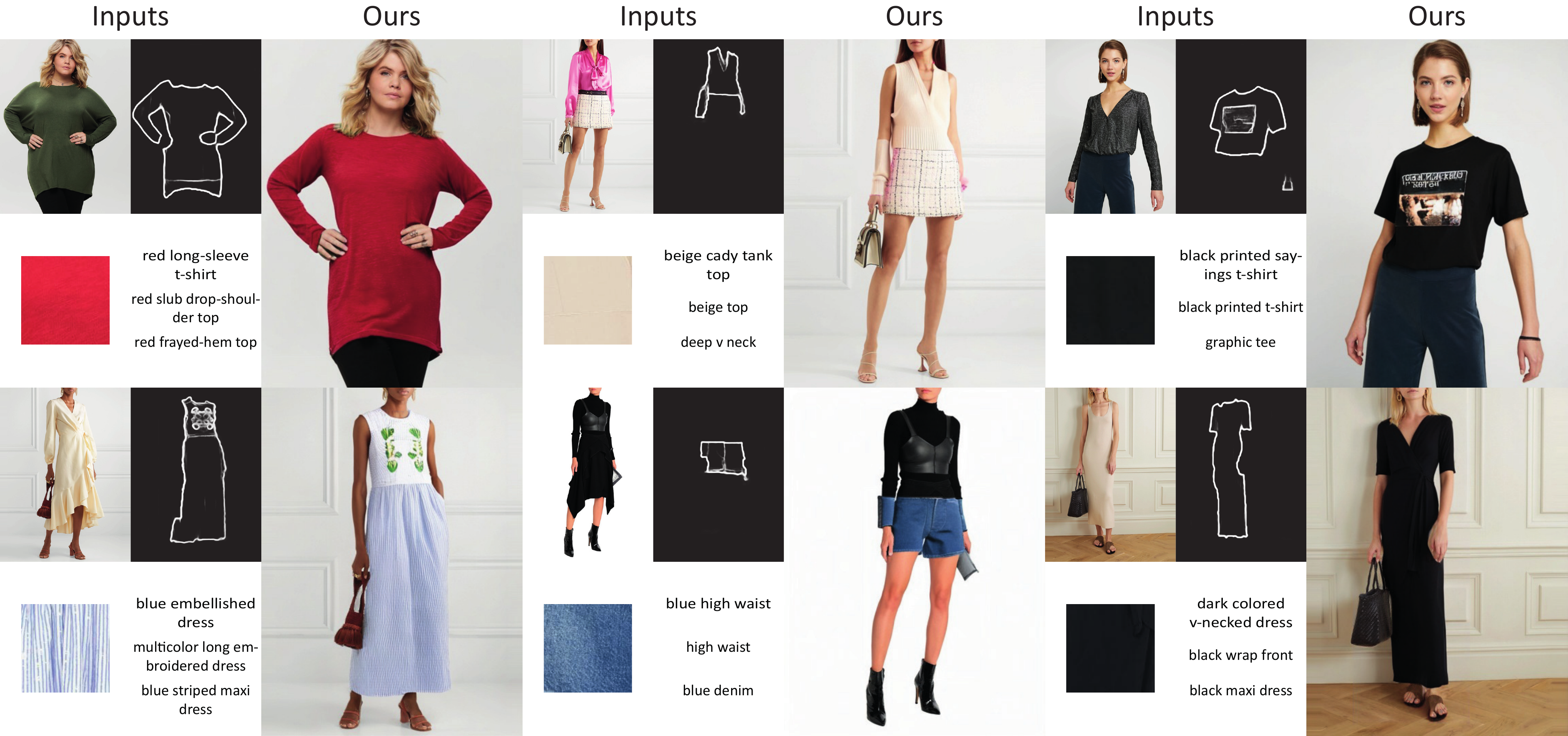}
\end{center}
\vspace{-0.2cm}
\caption{Failure cases of our proposed approach (\ours) on the \dataset and \datasetviton datasets.} 
\label{fig:qualitative-failure}
\vspace{-0.35cm}
\end{figure*}

\tit{Limitations and Failure Cases}
Fig.~\ref{fig:qualitative-failure} shows some failure cases of the proposed approach.  In the example in the first row and first column, we can see that the body shape of the generated model does not match the original one. We attribute this discrepancy to the inability of keypoints to describe body shape fully. In this sense, a potential future direction could involve augmenting the pose input with dense or 3D information. The image in the second row and first column demonstrates that when the sketch comprises distinct areas, each delineated by an edge, it could happen that only one delimited area of the generated garment is conditioned on texture. To solve this issue, a future direction could involve introducing spatial control for texture conditioning. Examples in the second column highlight the dependence of our model performance on the provided sketch. When the geometric warping module fails to generate a sketch able to fit the model's shape, the generation task also fails, resulting in unwanted artifacts. The examples presented in the third column reveal that our model sometimes fails to generate hands and text accurately when they occupy a limited area within the source image. This behavior is intrinsic in latent diffusion models~\cite{rombach2022high} and stems from the high compression nature of the latent space.

%% file: tables/model/main.tex
\begin{table*}[ht!]
\begin{center}
\caption{Quantitative results of \ours approach against competitors on the \dataset and \datasetviton datasets for both paired and unpaired settings. When considering evaluation metrics, PD stands for Pose Distance, SD for Sketch Distance, and TS for Texture Similarity. Best results are in bold, second best are underlined.}
\label{tab:main_merged}
\footnotesize
\setlength{\tabcolsep}{.35em}
\resizebox{\linewidth}{!}{
\begin{tabular}{lc cccc c cccccc c cccccc}
\toprule
 & & \multicolumn{4}{c}{\textbf{Modalities}} & & \multicolumn{6}{c}{\textbf{\dataset}}  & & \multicolumn{6}{c}{\textbf{\datasetviton}} \\
\cmidrule{3-6} \cmidrule{8-13} \cmidrule{15-20}
\textbf{Model} & & \textbf{Text} & \textbf{Pose} & \textbf{Sketch} & \textbf{Texture} & & \textbf{FID} $\downarrow$ & \textbf{KID} $\downarrow$ & \textbf{CLIP-S} $\uparrow$ & \textbf{PD} $\downarrow$ & \textbf{SD} $\downarrow$ & \textbf{TS} $\uparrow$ & & \textbf{FID} $\downarrow$ & \textbf{KID} $\downarrow$ & \textbf{CLIP-S} $\uparrow$ & \textbf{PD} $\downarrow$ & \textbf{SD} $\downarrow$ & \textbf{TS} $\uparrow$ \\
\midrule
\textit{Paired setting} \\
\hspace{0.4cm}Stable Diffusion v1.5~\cite{rombach2022high} &  & \cmark & & & & & 19.14 & 11.0 & 28.55 & 9.20 & 0.372 & 0.490 & & 16.14 & 5.38 & 29.58 & 11.04 & 0.405 & 0.507 \\
\hspace{0.4cm}ControlNet~\cite{zhang2023adding} &  & \cmark & \cmark & & & & 17.71 & 9.87 & 28.88 & 7.68 & 0.353 & 0.491 & & 16.65 & 5.73 & 29.46 & 8.10 & 0.387 & 0.509 \\
\hspace{0.4cm}SDEdit~\cite{meng2022sdedit} &  & \cmark & \cmark & \cmark & & & 7.17 & 2.97 & 30.19 & 5.47 & 0.263 & 0.545 & & 12.22 & 4.09 & 28.63 & 6.51 & 0.307 & 0.570 \\
\hspace{0.4cm}ControlNet~\cite{zhang2023adding} &  & \cmark & \cmark & \cmark & & & 24.40 & 15.39 & 27.88 & 7.85 & 0.354 & 0.476 & & 22.06 & 10.02 & 28.71 & 8.21 & 0.377 & 0.484 \\
\hspace{0.4cm}ControlNet~\cite{zhang2023adding}+IP-Adapter~\cite{ye2023ip} &  & \cmark & \cmark & \cmark & \cmark & & 16.02 & 7.79 & 28.22 & 8.12 & 0.363 & 0.545 & & 14.98 & 5.52 & 28.61 & 8.12 & 0.354 & 0.574 \\
\midrule
\hspace{0.4cm}MGD~\cite{baldrati2023multimodal} &  & \cmark & \cmark & \cmark & & & 5.74 & 2.11 & \bf31.68 & 4.72 & 0.188 & 0.571 & & 10.60 & 3.26 & \bf32.39 & 5.94 & 0.243 & 0.587 \\
\rowcolor{blond}
\hspace{0.4cm}\textbf{\ours (SDv1)} & & \cmark & \cmark & \cmark & \cmark & & \bf3.46 & \bf0.66 & \underline{31.26} & \bf4.34 & \underline{0.176} & \underline{0.578} & & \underline{6.14} & \underline{0.78} & 31.16 & \underline{4.72} & \bf0.179 & \underline{0.616} \\
\rowcolor{blond}
\hspace{0.4cm}\textbf{\ours (SDv2)} &  & \cmark & \cmark & \cmark & \cmark & & \underline{3.79} & \underline{0.99} & 31.01 & \bf4.34 & \bf0.172  & \bf0.595 & & \bf6.04 & \bf0.63 & \underline{31.30} & \bf4.67 & \underline{0.187} & \bf0.624 \\
\midrule
\textit{Unpaired setting} \\
\hspace{0.4cm}Stable Diffusion v1.5~\cite{rombach2022high} &  & \cmark & & & & & 21.77 & 12.9 & 27.15 & 10.00 & 0.492 & 0.476 & & 17.87 & 6.37 & 27.73 & 11.81 & 0.588 & 0.498 \\
\hspace{0.4cm}ControlNet~\cite{zhang2023adding} &  & \cmark & \cmark & & & & 20.16 & 11.62 & 27.60 & 8.39 & 0.469 & 0.481 & & 19.17 & 7.47 & 27.60 & 8.83 & 0.557 & 0.495 \\
\hspace{0.4cm}SDEdit~\cite{meng2022sdedit} &  & \cmark & \cmark & \cmark & & & 8.79 & 3.67 & 27.65 & \bf6.13 & 0.354 & 0.532 & & 15.14 & 5.99 & 24.95 & 7.10 & 0.446 & 0.559 \\
\hspace{0.4cm}ControlNet~\cite{zhang2023adding} &  & \cmark & \cmark & \cmark & & & 26.66 & 17.33 & 26.65 & 8.51 & 0.462 & 0.469 & & 23.84 & 11.92 & 26.93 & 8.88 & 0.547 & 0.480 \\
\hspace{0.4cm}ControlNet~\cite{zhang2023adding}+IP-Adapter~\cite{ye2023ip} &  & \cmark & \cmark & \cmark & \cmark & & 17.79 & 8.89 & 27.04 & 8.84 & 0.475 & 0.534  & & 17.73 & 7.25 & 26.64 & 8.84 & 0.507 & 0.561 \\
\midrule
\hspace{0.4cm}MGD~\cite{baldrati2023multimodal} &  & \cmark & \cmark & \cmark & & & 7.73 & 2.82 & \bf30.04 & 6.79 & 0.342 & 0.554 & & 12.81 & 3.86 & \bf30.75 & 7.22 & 0.331 & 0.578 \\
\rowcolor{blond}
\hspace{0.4cm}\textbf{\ours (SDv1)} &  & \cmark & \cmark & \cmark & \cmark & & \underline{5.69} & \underline{1.33} & 29.44 & \underline{6.19} & \underline{0.222} & \underline{0.577} & & \underline{10.18} & \underline{1.96} & 28.56 & \underline{6.59} & \bf0.239 & \underline{0.608} \\
\rowcolor{blond}
\hspace{0.4cm}\textbf{\ours (SDv2)} & & \cmark & \cmark & \cmark & \cmark & & \bf5.68 & \bf1.32 & \underline{29.78} & 6.26 & \bf0.218 & \bf0.597 &  & \bf9.30 & \bf1.26 & \underline{29.43} & \bf6.56 & \underline{0.247} & \bf0.630  \\
\bottomrule
\end{tabular}
}
\end{center}
\vspace{-0.4cm}
\end{table*}

%% file: tables/model/main_categories.tex
\begin{table*}[t]
\begin{center}
\caption{Category-wise quantitative results of \ours approach on the \dataset dataset for both paired and unpaired settings. When analyzing the modalities, T stands for Text, P for Pose, S for Sketch, and F for Fabric texture.}
\label{tab:dresscode_categories}
\footnotesize
\setlength{\tabcolsep}{.3em}
\resizebox{\linewidth}{!}{
\begin{tabular}{lc cccc c cccccc c cccccc c cccccc}
\toprule
 & & \multicolumn{4}{c}{\textbf{Modalities}} & & \multicolumn{6}{c}{\textbf{Upper-body}} & & \multicolumn{6}{c}{\textbf{Lower-body}} & & \multicolumn{6}{c}{\textbf{Dresses}} \\
\cmidrule{3-6} \cmidrule{8-13} \cmidrule{15-20} \cmidrule{22-27}
\textbf{Model} & & \textbf{T} & \textbf{P} & \textbf{S} & \textbf{F} & & \textbf{FID} $\downarrow$ & \textbf{KID}  $\downarrow$ & \textbf{CLIP-S} $\uparrow$ & \textbf{PD} $\downarrow$ & \textbf{SD} $\downarrow$ & \textbf{TS} $\uparrow$ & & \textbf{FID} $\downarrow$ & \textbf{KID} $\downarrow$ & \textbf{CLIP-S} $\uparrow$ & \textbf{PD} $\downarrow$ & \textbf{SD} $\downarrow$ & \textbf{TS} $\uparrow$ & & \textbf{FID} $\downarrow$ & \textbf{KID} $\downarrow$ & \textbf{CLIP-S} $\uparrow$ & \textbf{PD} $\downarrow$ & \textbf{SD} $\downarrow$ & \textbf{TS} $\uparrow$ \\
\midrule
\textit{Paired setting} \\
\hspace{0.4cm}Stable Diffusion v1.5~\cite{rombach2022high}  & & \cmark & & & & & 21.61 & 8.95 & 29.37 & 8.03 & 0.309 & 0.480 & & 29.37 & 15.29 & 27.67 & 9.64 & 0.358 & 0.496 &  & 37.83 & 22.51 & 28.59 & 9.88 & 0.449 & 0.493 \\
\hspace{0.4cm}ControlNet~\cite{zhang2023adding} & & \cmark & \cmark & & & & 21.51 & 8.97 & 29.36 & 6.32 & 0.287 & 0.479 & & 25.35 & 11.25 & 28.38 & 8.54 & 0.334 & 0.499 &  & 36.34 & 20.64 & 28.91 & 8.25 & 0.439 & 0.495 \\
\hspace{0.4cm}SDEdit~\cite{meng2022sdedit}  & & \cmark & \cmark & \cmark & & & 12.20 & 2.40 & 30.28 & 4.40 & 0.232 & 0.527 &  & 12.59 & 2.72 & 29.48 & 6.60 & 0.266 & 0.542 & & 16.48 & 6.09 & 30.81 & 5.63 & 0.291 & 0.565 \\
\hspace{0.4cm}ControlNet~\cite{zhang2023adding}  & & \cmark & \cmark & \cmark & & & 27.04 & 13.79 & 28.51 & 6.25 & 0.279 & 0.466 &  & 34.46 & 19.71 & 27.22 & 8.88 & 0.326 & 0.482 &  & 45.70 & 27.96 & 27.92 & 8.52 & 0.457 & 0.480 \\
\hspace{0.4cm}ControlNet~\cite{zhang2023adding}+IP-Adapter~\cite{ye2023ip}  & & \cmark & \cmark & \cmark & \cmark & & 18.10 & 6.25 & 28.71 & 6.42 & 0.291 & 0.544 &  & 21.95 & 8.72 & 27.61 & 8.90 & 0.337 & 0.543 &  & 38.98 & 20.72 & 28.34 & 9.07 & 0.462 & 0.548 \\
\midrule
\hspace{0.4cm}MGD~\cite{baldrati2023multimodal} & & \cmark & \cmark & \cmark & & & 12.42 & 3.71 & \bf31.90 & 3.72 & 0.180 & 0.547 & & 10.70 & 2.01 & \bf31.10 & 5.70 & 0.200 & 0.567 & & 11.38 & 1.89 & \underline{32.02} & 4.93 & 0.182 & 0.592 \\
\rowcolor{blond}
\hspace{0.4cm}\textbf{\ours (SDv1)} & & \cmark & \cmark & \cmark & \cmark & & \bf7.92 & \bf0.76 & \underline{31.07} & \underline{3.38} & \underline{0.161} & \underline{0.559} & & \bf7.22 & \bf0.59 & 30.60 & \underline{5.28} & \underline{0.192} & \underline{0.572} & & \bf9.14 & \bf0.92 & \bf32.10 & \bf4.52 & \underline{0.166} & \underline{0.620} \\
\rowcolor{blond}
\hspace{0.4cm}\textbf{\ours (SDv2)} & & \cmark & \cmark & \cmark & \cmark & & \underline{8.01} & \underline{0.97} & 30.78 & \bf3.33 & \bf0.160 & \bf0.568 & & \underline{7.56} & \underline{0.95} & \underline{30.66} & \bf5.25 & \bf0.191 & \bf0.582 & & \underline{9.69} & \underline{1.63} & 31.60 & \underline{4.57} & \bf0.164 & \bf0.637 \\
\midrule
\textit{Unpaired setting} \\
\hspace{0.4cm}Stable Diffusion v1.5~\cite{rombach2022high}  & & \cmark & & & & & 25.08 & 11.01 & 27.76 & 8.60 & 0.427 & 0.470 &  & 33.16 & 17.87 & 26.22 & 10.86 & 0.463 & 0.478 &  & 40.85 & 24.67 & 27.48 & 10.62 & 0.587 & 0.480 \\
\hspace{0.4cm}ControlNet~\cite{zhang2023adding} & & \cmark & \cmark & & & & 24.59 & 10.80 & 27.72 & 6.72 & 0.397 & 0.469 &  & 29.53 & 13.74 & 27.17 & 9.50 & 0.440 & 0.485 &  & 38.83 & 22.54 & 27.90 & 9.06 & 0.570 & 0.487 \\
\hspace{0.4cm}SDEdit~\cite{meng2022sdedit}  & & \cmark & \cmark & \cmark & & & 14.52 & 3.18 & 27.40 & 4.76 & 0.325 & 0.518 &  & 15.73 & 3.58 & 27.36 & 7.51 & 0.347 & 0.526 &  & 18.99 & 8.02 & 28.20 & 6.36 & 0.391 & 0.553 \\
\hspace{0.4cm}ControlNet~\cite{zhang2023adding} & & \cmark & \cmark & \cmark & & & 30.58 & 16.35 & 26.79 & 6.94 & 0.388 & 0.459 & & 37.98 & 22.35 & 26.12 & 9.50 & 0.420 & 0.473 & & 47.32 & 29.30 & 27.05 & 9.18 & 0.578 & 0.474 \\
\hspace{0.4cm}ControlNet~\cite{zhang2023adding}+IP-Adapter~\cite{ye2023ip}  & & \cmark & \cmark & \cmark & \cmark & & 21.32 & 7.96 & 26.98 & 7.03 & 0.396 & 0.528 & & 25.20 & 9.64 & 26.77 & 9.69 & 0.430 & 0.533 & & 40.33 & 22.34 & 27.37 & 9.84 & 0.541 & 0.599 \\
\midrule
\hspace{0.4cm}MGD~\cite{baldrati2023multimodal} & & \cmark & \cmark & \cmark & & & 15.99 & 4.50 & \bf29.76 & 5.41 & 0.327 & 0.532 & & 14.82 & 2.81 & \bf29.96 & 7.96 & 0.352 & 0.561 & & 14.71 & 3.63 & 30.41 & 7.15 & 0.348 & 0.568 \\
\rowcolor{blond}
\hspace{0.4cm}\textbf{\ours (SDv1)}& & \cmark & \cmark & \cmark & \cmark & & \underline{12.33} & \underline{1.71} & 28.50 & \bf4.59 & \underline{0.223} & \underline{0.555} & & \bf12.93 &  \bf1.51 & 29.34 & \bf7.52 & \underline{0.236} & \underline{0.566} & & \underline{12.65} & \bf1.96 & \underline{30.49} & \bf6.62 & \underline{0.208} & \underline{0.609} \\
\rowcolor{blond}
\hspace{0.4cm}\textbf{\ours (SDv2)}& & \cmark & \cmark & \cmark & \cmark & & \bf12.01 & \bf1.32 & \underline{29.08} & \underline{4.63} & \bf0.220 & \bf0.573 & & \underline{13.31} & \underline{1.90} & \underline{29.57} & \underline{7.55} & \bf0.231 & \bf0.582 & & \bf12.56 & \underline{2.02} & \bf{30.69} & \underline{6.72} & \bf0.203 & \bf0.635 \\
\bottomrule
\end{tabular}
}
\end{center}
\vspace{-0.4cm}
\end{table*}

%% file: tables/model/user_study.tex
\begin{table}[t]
\begin{center}
\caption{User study results on the unpaired setting of both \dataset and \datasetviton datasets. We report the percentage of times an image from \ours is preferred against a competitor. 
Note that when comparing against ControlNet with all modalities, we employ IP-Adapter to condition on texture.}
\label{tab:user_study}
\footnotesize
\setlength{\tabcolsep}{.32em}
\resizebox{\linewidth}{!}{
\begin{tabular}{cc cccc c ccc c ccc}
\toprule
& &\multicolumn{4}{c}{\textbf{Modalities}} & & \multicolumn{3}{c}{\textbf{Realism}} & & \multicolumn{3}{c}{\textbf{Multimodal Coherence}} \\
\cmidrule{3-6} \cmidrule{8-10} \cmidrule{12-14}
& & \textbf{T} & \textbf{P} & \textbf{S} & \textbf{F} & & 
 \textbf{SD} & \textbf{ControlNet} & \textbf{SDEdit} & & 
\textbf{SD} & \textbf{ControlNet} & \textbf{SDEdit} \\
\midrule
\multirow{4}{*}{{\rotatebox[origin=c]{90}{Dress}}} & \multirow{4}{*}{{\rotatebox[origin=c]{90}{Code M.}}} 
& \cmark & & & & & 94.54 & - & - & & 77.92 & - & - \\
& & \cmark & \cmark & & & & - & 91.89 & - & & - & 79.07 & - \\
& & \cmark & \cmark & \cmark & & & . & 96.10 & 80.21 & & - & 94.44 & 67.65 \\
& & \cmark & \cmark & \cmark & \cmark & & - & 97.67 & - & & - & 84.15 & - \\
\midrule
\multirow{4}{*}{{\rotatebox[origin=c]{90}{VITON}}} & \multirow{4}{*}{{\rotatebox[origin=c]{90}{HD M.}}} 
& \cmark & & & & & 95.83 & - & - & & 84.34 & - & - \\
& & \cmark & \cmark & & & & . & 95.60 & - & & - & 77.38 & - \\
& & \cmark & \cmark & \cmark & & & - & 94.25 & 64.94 & & - & 82.05 & 78.05 \\
& & \cmark & \cmark & \cmark & \cmark & & - & 96.62 & - & & - & 89.62 & - \\
\bottomrule
\end{tabular}
}
\end{center}
\vspace{-.4cm}
\end{table}

%% file: tables/model/modalities_ablation_unpaired.tex
\begin{table}[t]
\begin{center}
\caption{Performance analysis of our proposed model (\ours) on the unpaired setting of both \dataset and \datasetviton datasets as input modalities vary.}
\label{tab:modalities_unpaired}
\footnotesize
\setlength{\tabcolsep}{.4em}
\resizebox{\linewidth}{!}{
\begin{tabular}{cccc c cccccc}
\toprule
\multicolumn{4}{c}{\textbf{Modalities}} & & \multicolumn{5}{c}{\textbf{\dataset}} \\
\cmidrule{0-3} \cmidrule{6-11} 
\textbf{Text} & \textbf{Pose} & \textbf{Sketch} & \textbf{Texture} & & 
\textbf{FID} $\downarrow$ & \textbf{KID} $\downarrow$ & \textbf{CLIP-S} $\uparrow$  & \textbf{PD} $\downarrow$ & \textbf{SD} $\downarrow$ & \textbf{TS} $\uparrow$ \\
\midrule
\cmark & & & & & 6.43 & 1.35 & 30.41 & 7.24 & 0.404 & 0.539 \\
\cmark & \cmark & & & & 6.45 & 1.50 & 30.18 & 6.42 & 0.374 & 0.539 \\
\cmark & \cmark & \cmark & & & 6.53 & 1.87 & \bf30.44 & 6.28 & 0.225 & 0.553 \\
\rowcolor{blond}
\cmark & \cmark & \cmark & \cmark & & \bf5.68 & \bf1.32 & 29.78 & \bf6.26 & \bf0.218 & \bf0.597 \\
\midrule
\multicolumn{4}{c}{\textbf{Modalities}} & & \multicolumn{5}{c}{\textbf{\datasetviton}} \\
\cmidrule{0-3} \cmidrule{6-11} 
\textbf{Text} & \textbf{Pose} & \textbf{Sketch} & \textbf{Texture} & & 
\textbf{FID} $\downarrow$ & \textbf{KID} $\downarrow$ & \textbf{CLIP-S} $\uparrow$  & \textbf{PD} $\downarrow$ & \textbf{SD} $\downarrow$ & \textbf{TS} $\uparrow$ \\
\midrule
\cmark & & & & & 10.37 & 1.54 & 29.37 & 8.18 & 0.493 & 0.559 \\
\cmark & \cmark & & & & 10.53 & 1.71 & 29.31 & 7.26 & 0.472 & 0.560 \\
\cmark & \cmark & \cmark & & & 10.22 & 1.86 & \bf29.62 & 6.56 & 0.249 & 0.581 \\
\rowcolor{blond}
\cmark & \cmark & \cmark & \cmark & & \bf9.30 & \bf1.26 & 29.43 & \bf6.55 & \bf0.247 & \bf0.630 \\
\bottomrule
\end{tabular}
}
\end{center}
\vspace{-0.3cm}
\end{table}

%% file: tables/model/uncond.tex
\begin{table}[t]
\begin{center}
\caption{Ablation study of our complete model varying the unconditional portion during training and the sketch conditioning rate at inference time. Results refer to the unpaired setting.}
\label{tab:dressCode_ablations}
\footnotesize
\setlength{\tabcolsep}{.35em}
\resizebox{\linewidth}{!}{
\begin{tabular}{cc c cccccc}
\toprule
& & & \multicolumn{6}{c}{\textbf{\dataset}} \\
\cmidrule{4-9}
\textbf{Uncond. Portion} & \textbf{Sketch Cond.} & & 
\textbf{FID} $\downarrow$ & \textbf{KID} $\downarrow$ & \textbf{CLIP-S} $\uparrow$  & \textbf{PD} $\downarrow$ & \textbf{SD} $\downarrow$ & \textbf{TS} $\uparrow$\\
\midrule
0.1 & 1.0 & & 9.91 & 4.94 & 27.15 & 6.42 & 0.148 & 0.576 \\
0.2 & 1.0 & & 9.64 & 4.55 & 27.04 & 6.48 & 0.155 & 0.576 \\
0.3 & 1.0 & & 9.95 & 4.77 & 27.25 & 6.49 & 0.164 & 0.573 \\
\midrule
0.2 & 0.8 & & 8.75 & 3.81 & 27.61 & 6.48 & 0.166 & 0.580 \\
0.2 & 0.6 & & 7.91 & 3.03 & 28.10 & 6.44 & 0.175 & 0.586 \\
0.2 & 0.4 & & 6.56 & 1.87 & 28.76 & 6.35 & 0.182 & 0.593\\
\rowcolor{blond}
0.2 & 0.2 & & 5.68 & 1.32 & 29.78 & 6.26 & 0.218 & 0.597 \\
0.2 & 0.0 & & 5.64 & 1.19 & 29.30 & 6.40 & 0.370 & 0.584 \\
\bottomrule
\end{tabular}
}
\end{center}
\vspace{-0.4cm}
\end{table}

%% file: tables/model/cross_attention.tex
        
\begin{table}[t]
\begin{center}
\caption{Quantitative results of our proposed approach (\ours) on \dataset dataset when using different cross-attention groups for conditioning on texture and text input. For a visual representation of the cross-attention groups, see Fig.~\ref{fig:cross_attention_groups}.
Here, \tcbox[on line, boxsep=2pt, left=-1.2pt,right=-1.2pt,top=0pt,bottom=-0.5pt, sharp corners,
        colframe=white,colback=cross-attn-blue]{\textcolor{cross-attn-blue}{a}} stands for the cross-attention groups conditioned on the texture, while \tcbox[on line, boxsep=2pt, left=-1.2pt,right=-1.2pt,top=0pt,bottom=-0.5pt, sharp corners,
        colframe=white,colback=cross-attn-yellow]{\textcolor{cross-attn-yellow}{a}} are cross-attention groups conditioned on the text.}
\label{tab:cross-annt-ablation}
\footnotesize
\setlength{\tabcolsep}{.4em}
\resizebox{0.9\linewidth}{!}{
\begin{tabular}{c c cccccc}
\toprule
\textbf{Cross-Attn.} & & \multicolumn{6}{c}{\textbf{\dataset}} \\
\cmidrule{1-1} \cmidrule{3-8}
\textbf{Groups} & & 
\textbf{FID} $\downarrow$ & \textbf{KID} $\downarrow$ & \textbf{CLIP-S} $\uparrow$  & \textbf{PD} $\downarrow$ & \textbf{SD} $\downarrow$ & \textbf{TS} $\uparrow$ \\
\midrule
\begin{minipage}{.1\textwidth}
      \includegraphics[width=\linewidth, height=2mm]{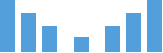}
    \end{minipage} & & 5.63 & 1.36 & 27.13 & 6.23 & 0.215 & 0.606 \\
\begin{minipage}{.1\textwidth}
      \includegraphics[width=\linewidth, height=2mm]{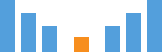}
    \end{minipage} & & 5.63 & 1.36 & 27.13 & 6.25 & 0.215 & 0.605 \\
\rowcolor{blond}
\begin{minipage}{.1\textwidth}
      \includegraphics[width=\linewidth, height=2mm]
      {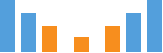}
    \end{minipage} & & 5.68 & 1.32 & 29.78 & 6.26 & 0.218 & 0.597 \\
\begin{minipage}{.1\textwidth}
      \includegraphics[width=\linewidth, height=2mm]{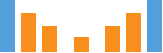}
    \end{minipage} & & 6.31 & 1.77 & 30.45 & 6.25 & 0.223 & 0.560 \\
\begin{minipage}{.1\textwidth}
      \includegraphics[width=\linewidth, height=2mm]{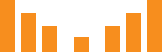}
    \end{minipage} & & 6.53 & 1.87 & 30.44 & 6.28 & 0.225 & 0.553 \\
\midrule
\begin{minipage}{.1\textwidth}
      \includegraphics[width=\linewidth, height=2mm]{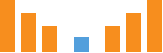}
    \end{minipage} & & 6.53 & 1.87 & 30.44 & 6.27 & 0.225 & 0.553 \\
\begin{minipage}{.1\textwidth}
      \includegraphics[width=\linewidth, height=2mm]{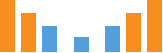}
    \end{minipage} & & 6.22 & 1.77 & 28.15 & 6.21 & 0.217 & 0.572 \\
\begin{minipage}{.1\textwidth}
      \includegraphics[width=\linewidth, height=2mm]{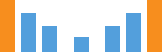}
    \end{minipage} & & 5.62 & 1.38 & 27.22 & 6.22 & 0.215 & 0.604 \\
\bottomrule
\end{tabular}
}
\end{center}
\vspace{-0.4cm}
\end{table}

%% file: sections/06_conclusion.tex
In this work, we have presented \ours, a novel approach for multimodal-conditioned fashion image editing in which the image generation process is conditioned with multiple modalities, such as text, body pose, garment sketch, and fabric texture. This is achieved by extending latent diffusion models to incorporate these different modalities and modifying the structure of the denoising network to take multimodal inputs. To effectively incorporate texture information, we have leveraged textual inversion techniques and proposed to combine text and texture features through the cross-attention layers of the denoising network. The proposed texture conditioning method enables fine-grained control over the generated images without adding denoising network parameters. To validate our approach, we have also extended two existing fashion datasets with multimodal annotations using a semi-automatic procedure.
Our comprehensive experiments on standard and newly proposed metrics validate the effectiveness of \ours, outperforming state-of-the-art methods in terms of realism and coherence with multimodal inputs. Conclusively, the proposed approach not only sets a new standard for multimodal-conditioned fashion image editing but also opens avenues for further research at the intersection of computer vision and fashion. The presented results represent one of the first successful attempts to imitate the work of designers in the creative process of fashion design and could be a starting point for the widespread adoption of diffusion models in the creative industries, oversight by human input.